\documentclass[10pt,twocolumn,letterpaper]{article}

\usepackage[pagenumbers]{cvpr}              

\definecolor{sns_blue}{rgb}{0.21, 0.06, 0.42}    
\definecolor{sns_violet}{rgb}{0.45, 0.12, 0.51}  
\definecolor{sns_orange}{rgb}{0.75, 0.22, 0.46}  

\newcommand{\noIndentHeading}[1]{\noindent\textbf{#1}}

\newcommand{\methodName}{SPOOF\xspace}

\newcommand{\dnns}{DNNs\xspace}

\newcommand{\alexNet}{AlexNet\xspace}

\newcommand{\resNetFifty}{ResNet-50\xspace}

\newcommand{\vitB}{ViT-B/16\xspace}

\newcommand{\imageNet}{ImageNet\xspace}
\newcommand{\imageNetOneK}{ImageNet-1K\xspace}

\newcommand{\mnist}{MNIST\xspace}
\newcommand{\val}{Val\xspace}

\newcommand{\cluneRawFool}{Direct-Fool\xspace}
\newcommand{\cluneCPPNFool}{CPPN-Fool\xspace}
\newcommand{\sfi}{SFI\xspace}

\newcommand{\pcr}{PCR\xspace}

\definecolor{cvprblue}{rgb}{0.21,0.49,0.74}
\usepackage[pagebackref,breaklinks,colorlinks,allcolors=cvprblue]{hyperref}

\usepackage{algorithm}
\usepackage{algorithmic}

\usepackage{booktabs}
\usepackage{multirow}
\usepackage{siunitx}
\usepackage{array}

\def\paperID{19726} 
\def\confName{CVPR}
\def\confYear{2026}

\title{\methodName: Simple Pixel Operations for Out-of-Distribution Fooling}

\author{
Ankit Gupta$^{1,2,*}$\quad
Christoph Adami$^{2,3,4}$\quad
Emily Dolson$^{1,2}$\\[6pt]
\parbox{\linewidth}{\centering
$^{1}$Department of Computer Science \& Engineering \quad
$^{2}$Program in Evolution, Ecology, \& Behavior\\
$^{3}$Department of Physics \& Astronomy \quad
$^{4}$Department of Microbiology, Genetics, \& Immunology \quad
Michigan State University
}\\[6pt]
{\tt\small guptaa23@msu.edu, adami@msu.edu, dolsonem@msu.edu}\\[-2pt]
{\small $^{*}$Corresponding author}
}

\begin{document}
\twocolumn[{%
\maketitle

 \vspace{-1cm} 

\begin{center}
\begin{minipage}{1.0\linewidth}
    \centering
    \renewcommand{\arraystretch}{1.1} 
    \setlength{\tabcolsep}{6pt} 
    \begin{tabular}{|c|c|}
        \hline
        \textbf{Class 84: Peacock | SPOOF Progress | ViT-B/16} & \textbf{Class 640: Manhole Cover | All Attacks | Models} \\
        \hline
        \includegraphics[width=0.48\linewidth]{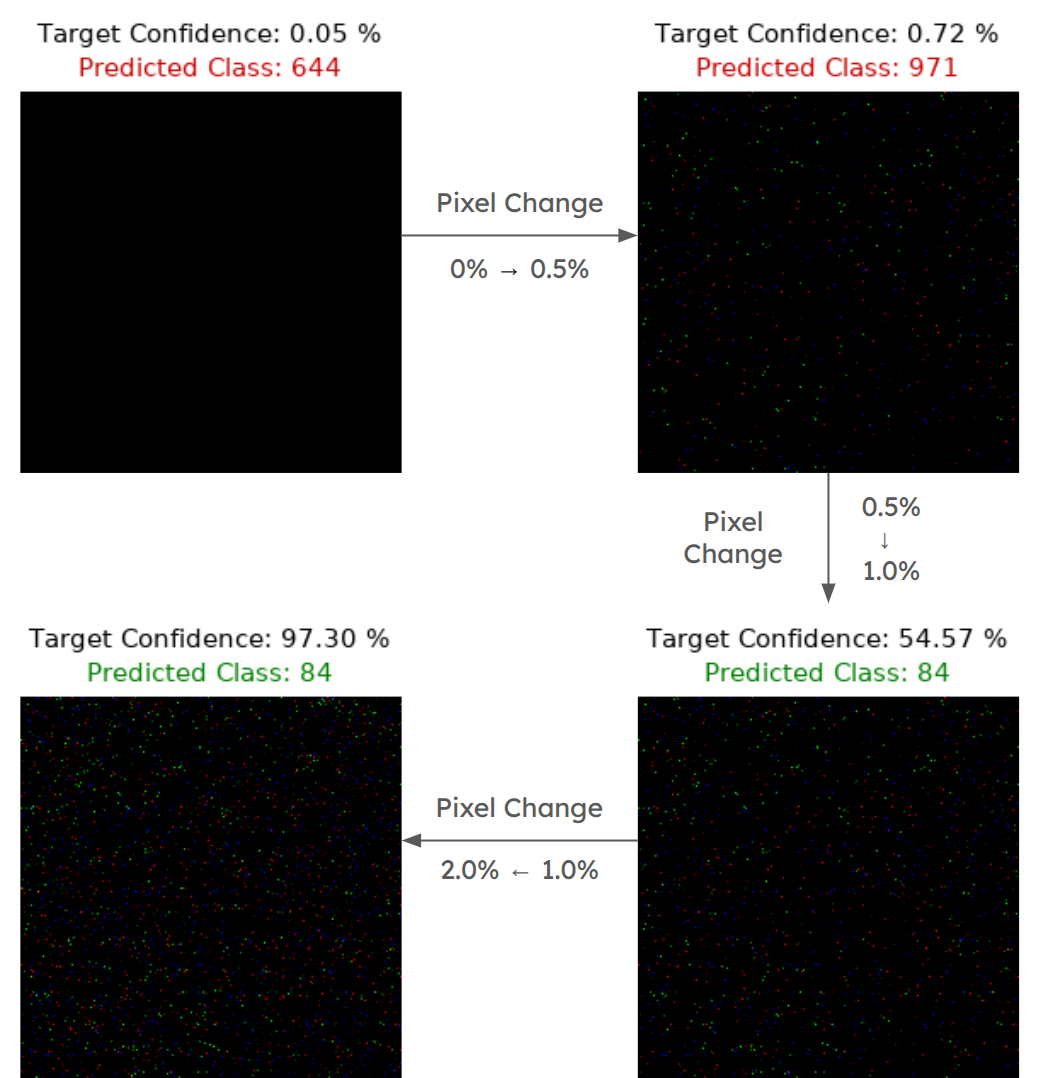} &
        \includegraphics[width=0.48\linewidth]{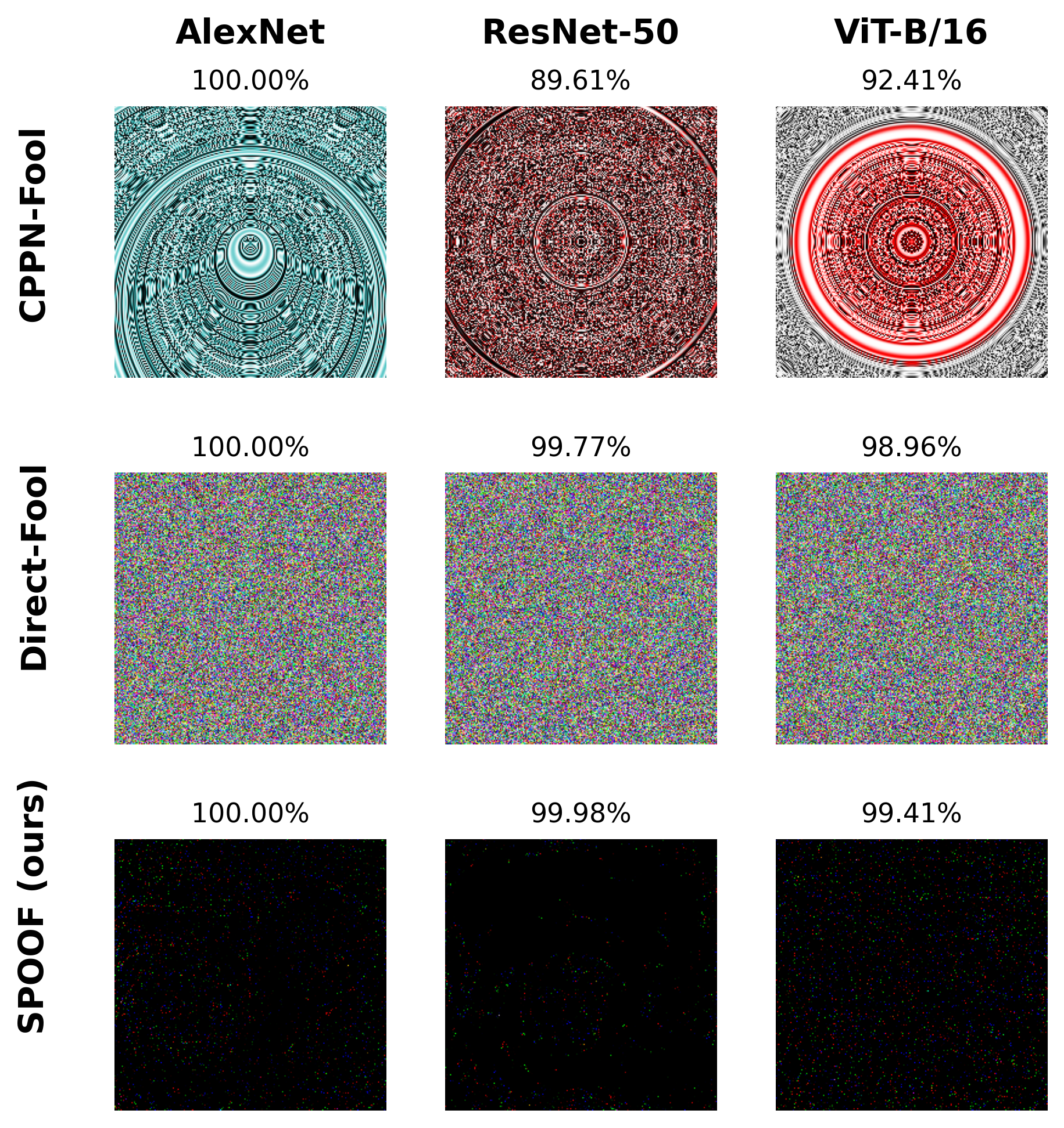} \\
        \hline
    \end{tabular}


    \captionof{figure}{
    \textbf{(Left)} \methodName applied to a blank canvas on an ImageNet-trained \textbf{\vitB}. Clockwise images show a steady rise in target-class 84 confidence as more pixels are retained (0-2\%) through \methodName's greedy updates. Predicted-class text appears in \textcolor{green}{green} when the model's top-1 prediction matches the target class and in \textcolor{red}{red} otherwise. 
    \textbf{(Right)} Fooling images for class~640 produced by three fooling approaches across three ImageNet classifiers. While CPPN-based fooling produces structured patterns and direct encoding yields dense noise, \methodName produces high-confidence fooling images using only extremely sparse, non-semantic pixel updates.
    }
    \label{fig:poster}

\end{minipage}
\end{center}

\vspace{.5em} 
\thispagestyle{empty}

 \begin{abstract}
Deep neural networks (DNNs) excel across image recognition tasks, yet continue to exhibit overconfidence on inputs that bear no resemblance to natural images. Revisiting the “fooling images” work introduced by Nguyen et al. (2015), we re-implement both CPPN-based and direct-encoding-based evolutionary fooling attacks on modern architectures, including convolutional and transformer classifiers. Our re-implementation confirm that high-confidence fooling persists even in state-of-the-art networks, with transformer-based ViT-B/16 emerging as the most susceptible—achieving near-certain misclassifications with substantially fewer queries than convolution-based models. We then introduce SPOOF, a minimalist, consistent, and more efficient black-box attack generating high-confidence fooling images. Despite its simplicity, SPOOF generates unrecognizable fooling images with minimal pixel modifications and drastically reduced compute. Furthermore, retraining with fooling images as an additional class provides only partial resistance, as SPOOF continues to fool consistently with slightly higher query budgets—highlighting persistent fragility of modern deep classifiers.
\end{abstract}
 }]

\section{Introduction}
    Deep neural networks (\dnns) have achieved phenomenal success in vision tasks such as image classification, detection and segmentation \cite{krizhevsky2012imagenet}, sometimes surpassing even human-level accuracy \cite{he2015delving}. Yet, these classifiers can be surprisingly brittle---inputs altered by often imperceptible perturbations can lead to confidently incorrect predictions \cite{goodfellow2014explaining, guo2019simple}. Such vulnerabilities, known as adversarial attacks, involve subtle modifications to natural images. 
    
    While adversarial attacks focus on perturbing natural samples within a dataset, a complementary and underexplored direction examines the striking tendency of DNNs to assign high confidence to entirely unrecognizable, synthetic images that appear as random or patterned noise to humans. These were introduced as ``fooling images''by \cite{nguyen2015deep}, who showed ImageNet-trained \alexNet could yield near-certain predictions on images evolved via Compositional Pattern-Producing Networks (CPPNs) \cite{stanley2007compositional}. Intriguingly, when attempting direct pixel-space evolution (directly encoded images), they found it difficult to achieve the same level of high confidence fooling. Both types of fooling images were produced in a fully black-box setting using an evolutionary algorithm called Map-Elites \cite{Mouret_Clune_2015}.

    In this work, we begin by re-implementing the CPPN-based and direct pixel-encoding fooling attacks of \cite{nguyen2015deep} and evaluating them on both convolutional and transformer-based architectures. After reproducing the results on \imageNetOneK trained \alexNet, we extended the fooling attack to \resNetFifty and \vitB. Our re-implementation shows that all classifiers remain vulnerable, but their susceptibility differs substantially: \vitB is fooled the fastest and with the fewest queries---specially under direct pixel encoding---whereas \resNetFifty is consistently the most difficult to fool under both attack types (see \cref{fig:o_cppn_fool_plot,fig:e_raw_fool_plot}).
    
    Building on these observations, we introduce a much simpler fooling attack that relies on minimal pixel modifications yet delivers comparable or even stronger fooling performance while operating at orders-of-magnitude lower computational cost. We refer to this approach as \textbf{\methodName} (\textbf{S}imple \textbf{P}ixel \textbf{O}perations for \textbf{O}ut-of-Distribution \textbf{F}ooling), a fast and remarkably effective black-box method for producing unrecognizable fooling images. 
    
    Despite its algorithmic simplicity, being just a greedy hill climber at its core, \methodName consistently achieves high-confidence fooling. Starting from a blank image, \methodName updates one randomly selected pixel at a time, accepting a modification only when it increases the target-class confidence. Even with extremely sparse edits, this greedy procedure often drives predictions towards near-certainty, highlighting substantial vulnerability in both convolutional and transformer-based \imageNet classifiers (see \cref{fig:poster} for sample fooling images produced by \methodName across classifiers).

    Fooling attacks expose a failure mode that is, in some respects, more severe than that revealed by adversarial examples. While adversarial perturbations corrupt an otherwise natural input, fooling attacks elicit high-confidence predictions from images that are essentially empty---images that rarely contain class-relevant structure at any point. This reveals that the classifier's decision surface assigns confident semantics deep inside regions of the input space that have no connection to natural images. We study this vulnerability by revisiting the CPPN-based and direct-encoding attacks of \cite{nguyen2015deep} and by introducing \methodName, a sparse-pixel fooling attack analogous to sparse adversarial methods \cite{modas2019sparsefool, croce2019sparse}. \methodName makes the issue even more pronounced: not only are the images unrecognizable, they contain almost no active pixels at all.

    \section{Key Contributions}
    In a nutshell, this paper makes the following contributions:
    \begin{itemize}
        \item \textbf{A re-implementation and extension of \cite{nguyen2015deep}.}  
        We re-implement and reproduce the CPPN- and direct-encoding fooling attacks of \cite{nguyen2015deep}, applying them to modern classifiers and evaluating them under both the original and expanded experimental conditions.
   
        \item \textbf{SPOOF: a simple and efficient black-box fooling algorithm.}  
        We introduce \methodName, a greedy hill-climbing method with random pixel search that consistently generates high-confidence, unrecognizable fooling images using minimal pixel modifications and substantially less compute.
        
        \item \textbf{\vitB{} is the easiest classifier to fool.}  
        Across re-implementation of directly-encoded fooling and \methodName, \vitB---arguably the current state-of-the-art classifier---consistently suffers high-confidence misclassifications with substantially fewer queries than convolution-based classifiers, highlighting its striking vulnerability to fit purely non-semantic noise, suggesting that its search space contains an abundance of high-confidence solutions embedded entirely within noise-dominated regions.

        \item \textbf{Fooling attacks are resilient to retraining.}  
        Fine-tuning with fooling images sometimes delays but rarely prevents fooling attacks; \methodName{} consistently regains high-confidence fooling under extended query budgets.

        \item \textbf{Fooling solutions are abundantly present even in full pixel-space noise.}
        Despite the vast dimensionality of the pixel space, fully random pixel-wise noise—-mirroring the direct-encoding setup of \cite{nguyen2015deep} and the random-initialization ablation of \methodName---still enables black-box search to reach high-confidence misclassifications. The ease with which these solutions emerge highlights how permissive the decision landscape is in noise-dominated regions. Among these noise-initialized images, ResNet-50 exhibits the strongest resistance, whereas ViT-B/16 is the easiest to fool.
        
    \end{itemize}

\section{Literature Review}
    The remarkable success of \dnns on large-scale image classification benchmarks such as \imageNetOneK~\cite{krizhevsky2012imagenet} was quickly followed by foundational results that exposed their susceptibility to carefully constructed perturbations \cite{szegedy2013intriguing, goodfellow2014explaining}. These early findings established adversarial vulnerability as a core limitation of modern classifiers and gave rise to an extensive body of work on adversarial attacks under various threat models and constraints. 

    \noIndentHeading{Adversarial Attacks on Natural Images.}
    Most adversarial attacks focus on natural images, introducing subtle perturbations that alter classifier predictions while keeping the input visually unchanged. These attacks are typically categorized by the information available to the adversary. In white-box settings, full access to classifier parameters and gradients is assumed~\cite{goodfellow2014explaining, moosavi2016deepfool, madry2017towards, carlini2017towards, croce2020reliable}. Black-box attacks, by contrast, rely only on classifier outputs, making them more restrictive and realistic for deployed systems~\cite{ilyas2018black, brendel2018decision, chen2020hopskipjumpattack}. They can be further divided into \emph{score-based} (access to softmax outputs)~\cite{ilyas2018prior, li2019nattack, andriushchenko2020square} and \emph{label-based} (access to top-1 predictions) settings~\cite{li2020qeba}.
    
    Other key distinctions include \emph{targeted vs. non-targeted} attacks~\cite{kurakin2017adversarial}, \emph{transfer-based (surrogate) vs. query-based} methods~\cite{dong2019evading, papernot2017practical}, and constraints on perturbation size or pixel-level sparsity \cite{modas2019sparsefool, croce2019sparse, su2019onepixel, narodytska2016simple, bhagoji2018practical}. These approaches typically preserve image structure, yielding adversarial inputs that remain semantically recognizable.
    
    \noIndentHeading{Fooling Attacks from Unrecognizable Images.}
    While most conventional adversarial attacks operate on recognizable natural images that are modified imperceptibly to preserve semantics, another line of work explores the opposite extreme: crafting inputs that are entirely unrecognizable to humans, yet are confidently misclassified. 
    
    Nguyen \etal~\cite{nguyen2015deep} first demonstrated this phenomenon by evolving abstract, unrecognizable patterns that state-of-the-art ImageNet classifiers labeled with near-perfect confidence. More recently, \cite{kumano2023sparse} introduced ``Sparse Fooling Images" (\sfi), a raw‑pixel method based on differential evolution that perturbs only a few pixels on a blank canvas rather than the full image. The method works for targeted fooling on smaller datasets where very low pixel budgets can suffice. However, it failed to produce successful fooling images on \imageNet. 

    Work in this direction, most notably by Nguyen et al. \cite{nguyen2015deep} on \imageNet trained DNNs, indicates that state-of-the-art classifiers can fail in ways that lie entirely outside the scope of conventional adversarial perturbations. While out-of-distribution detection \cite{Liu_Wang_Owens_Li_2020, Sun_Guo_Li_2021, Sun_Ming_Zhu_Li_2022} and adversarial robustness have both been studied extensively, the behavior of modern models on deliberately unrecognizable, non-semantic inputs remains largely unexplored, despite constituting a close parallel to adversarial vulnerabilities. \methodName introduces an extreme instance of this setting, offering a complementary framework for probing vulnerabilities that traditional adversarial analysis does not capture.

\begin{table}[ht]
\centering
\caption{Parameter settings for fooling-attack experiments.}
\renewcommand{\arraystretch}{1.0}
\begin{tabular}{lccc}
\toprule
\textbf{Experiment} &
\multicolumn{1}{c}{\textbf{Population}} &
\multicolumn{1}{c}{\textbf{Generations}} &
\multicolumn{1}{c}{\textbf{Queries}} \\
\textbf{} &
\multicolumn{1}{c}{\textbf{Size}} &
\multicolumn{1}{c}{\textbf{(×1000)}} &
\multicolumn{1}{c}{\textbf{(×1000)}} \\
\midrule
\cluneCPPNFool \\ (Original)  & 400  & 5  & 2  \\
\specialrule{0.4pt}{0pt}{0pt}
\cluneCPPNFool \\ (Extended)  & 1000 & 50 & 50 \\
\specialrule{0.4pt}{0pt}{0pt}
\cluneRawFool \\ (Original)   & 400  & 20 & 8  \\
\specialrule{0.4pt}{0pt}{0pt}
\cluneRawFool \\ (Extended)   & 1000 & 50 & 50 \\
\specialrule{0.4pt}{0pt}{0pt}
SPOOF (on \\ pre-trained \\ classifiers)          & --   & 50 & 50 \\
\specialrule{0.4pt}{0pt}{0pt}
SPOOF (on \\ retrained \\ classifiers)          & --   & 150 & 150 \\

\bottomrule
\end{tabular}
\label{tab:parameters}
\end{table}

\section{Notations and Preliminaries}
        
    \noIndentHeading{Fooling Attack.}
    Analogous to adversarial attacks that perturb natural images to cause misclassification, we consider the complementary setting: synthesizing non-sensical, unrecognizable images that provoke high-confidence predictions for a chosen target. Nguyen et~al.~\cite{nguyen2015deep} introduced these as \emph{fooling images}; here we formally refer to the process of generating such fooling images as a \emph{fooling attack}. In this work we present a novel instance of such an attack, \methodName{}.

    \noIndentHeading{Re-implementation Experiments: \cluneRawFool and \cluneCPPNFool.}
    Nguyen et~al.~\cite{nguyen2015deep} introduced two evolutionary fooling attacks based on distinct image encodings. The first, a \emph{direct encoding}, represents each pixel’s value explicitly and evolves it through independent random mutations. The second, an \emph{indirect encoding}, employs Compositional Pattern Producing Networks (CPPNs)---neural networks that map pixel coordinates to color values, thereby enabling the emergence of structured, regular patterns such as symmetry and repetition \cite{stanley2007compositional}. In this work, we refer to their fooling attack experiment using directly encoded version as \emph{\cluneRawFool} and the indirectly encoded, CPPN-based version as \emph{\cluneCPPNFool}.

    \noIndentHeading{Experiment Parameters.}
    \cref{tab:parameters} lists the query budgets for our re-implementation of \cite{nguyen2015deep} and for \methodName. Since \cite{nguyen2015deep} employed Map-Elites as the evolutionary algorithm, the average query count per target corresponds to the total evaluations (\textit{population size} × \textit{generations}) divided by the number of ImageNet classes. In contrast, \methodName uses a single-sample hill-climber, where each iteration constitutes one query. For consistency with the pretrained runs, retraining experiments followed the same query budgets as their pretrained counterparts. The one exception is \methodName; because its fooling-confidence was still improving at the original query budget, we allowed an extended query budget so that we could determine at what point it reached convergence.
        
    \noIndentHeading{Evaluation Metrics.}
    All fooling-attack experiments were evaluated using the following metrics. Resulting metrics for each fooling experiment are aggregated over five random seeds and summarized as the \emph{median} across 1000 target classes, following \cite{nguyen2015deep}; mean results are reported in the supplementary material.
    
    \begin{itemize}

        \item \textit{Runtime:} Wall-clock time per fooling experiment on NVIDIA H200 GPUs, indicating the computational efficiency of the fooling attack, analogous to adversarial benchmarks for attack feasibility \cite{Croce2020RobustBench,Cina2024AttackBench}.

        \item \textit{Confidence:} Predicted probability assigned to the target label on the evaluated image, used as a standard indicator of attack strength across both fooling and adversarial benchmarks \cite{nguyen2015deep,kumano2023sparse,carlini2017towards,Croce2020RobustBench}.
                
        \item \textit{Fooling-ASR:} Introduced in this work for fooling attacks as an analog to the standard Attack Success Rate (ASR) used in adversarial benchmarks, Fooling-ASR measures the fraction of target classes successfully fooled within the query budget, offering a direct and interpretable indicator of fooling effectiveness \cite{carlini2017towards,ilyas2018black,andriushchenko2020square}.

        \item \textit{Pixel Change Ratio (\pcr):} Measures the proportion of pixels altered relative to the initial synthesized fooling image for a given target class, which is analogous to perturbation sparsity metrics in the adversarial attack literature~\cite{su2019onepixel,modas2019sparsefool,croce2019sparse}. For fooling attacks initialized from a blank canvas, this value corresponds to the fraction of non-zero pixels in the final image, revealing the surprising vulnerability of classifiers that confidently label near-empty images.

        \item \textit{Queries:} Average number of classifier queries per target (within the allowed budget), reporting attack query cost in the same spirit as query-limited adversarial evaluations~\cite{ilyas2018black,andriushchenko2020square}.

    \end{itemize}

\begin{table*}[ht]
\centering
\small
\setlength{\tabcolsep}{6pt}
\renewcommand{\arraystretch}{1.0}
\caption{Comparison of fooling performance across \imageNetOneK pre-trained classifiers on three benchmarks. SPOOF generally achieves comparable or higher confidence with far fewer pixel changes and lower runtime. Pixel Change Ratio (\pcr) is 100 \% for the re-implementation experiments of \cite{nguyen2015deep} as for \cluneCPPNFool the initial images are entirely replaced with new patterned images and for \cluneRawFool pixels are mutated for several generations leading to complete pixel change of initial image.}
\begin{tabular}{l l c c c c}
\toprule
\textbf{Classifier} & \textbf{Fooling Attack} &
\textbf{Runtime (hours)} &
\textbf{Confidence (\%)} &
\textbf{Fooling-ASR (\%)} &
\textbf{\pcr \%} \\ 
\midrule

\multirow{5}{*}{AlexNet}
 & \cluneCPPNFool (Original) & $2.81$ & $90.81^{\dagger}$ & $93.44$ & $100$ \\
 & \cluneCPPNFool (Extended) & $73.80$ & $99.15$ & $98.72$ & $100$ \\
 & \cluneRawFool (Original)  & $0.42$ & $25.89^{\dagger}$ & $65.40$ & $100$ \\
 & \cluneRawFool (Extended)  & $1.92$ & $\mathbf{99.91}$ & $\mathbf{99.14}$ & $100$ \\
 & \textbf{SPOOF (ours)} & $\mathbf{0.20}$ & $96.85$ & $89.12$ & $\mathbf{2.35}$ \\ 
\midrule

\multirow{5}{*}{ResNet-50}
 & \cluneCPPNFool (Original) & $2.84$ & $51.62$ & $87.00$ & $100$ \\
 & \cluneCPPNFool (Extended) & $75.28$ & $81.82$ & $\mathbf{97.12}$ & $100$ \\
 & \cluneRawFool (Original)  & $0.69$ & $28.67$ & $81.34$ & $100$ \\
 & \cluneRawFool (Extended)  & $3.06$ & $79.51$ & $96.44$ & $100$ \\
 & \textbf{SPOOF (ours)} & $\mathbf{0.58}$ & $\mathbf{98.02}$ & $96.04$ & $\mathbf{1.25}$ \\ 
\midrule

\multirow{5}{*}{ViT-B/16}
 & \cluneCPPNFool (Original) & $2.83$ & $79.48$ & $93.94$ & $100$ \\
 & \cluneCPPNFool (Extended) & $77.80$   & $92.41$   & $99.68$   & $100$ \\
 & \cluneRawFool (Original)  & $\mathbf{1.53}$ & $95.08$ & $\mathbf{100.00}$ & $100$ \\
 & \cluneRawFool (Extended)  & $6.39$ & $99.49$ & $\mathbf{100.00}$ & $100$ \\
 & \textbf{SPOOF (ours)} & $1.65$ & $\mathbf{99.61}$ & $\mathbf{100.00}$ & $\mathbf{3.62}$ \\

\bottomrule
\end{tabular}

\begin{flushleft}
{\footnotesize ${}^{\dagger}$These reproduced confidence scores closely match those reported by \cite{nguyen2015deep} (88.11\% and 21.59\%), confirming alignment with the original results.}
\end{flushleft}

\label{tab:spoof_results}
\end{table*}


\begin{figure}[ht]
    \centering
    \includegraphics[width=\columnwidth]{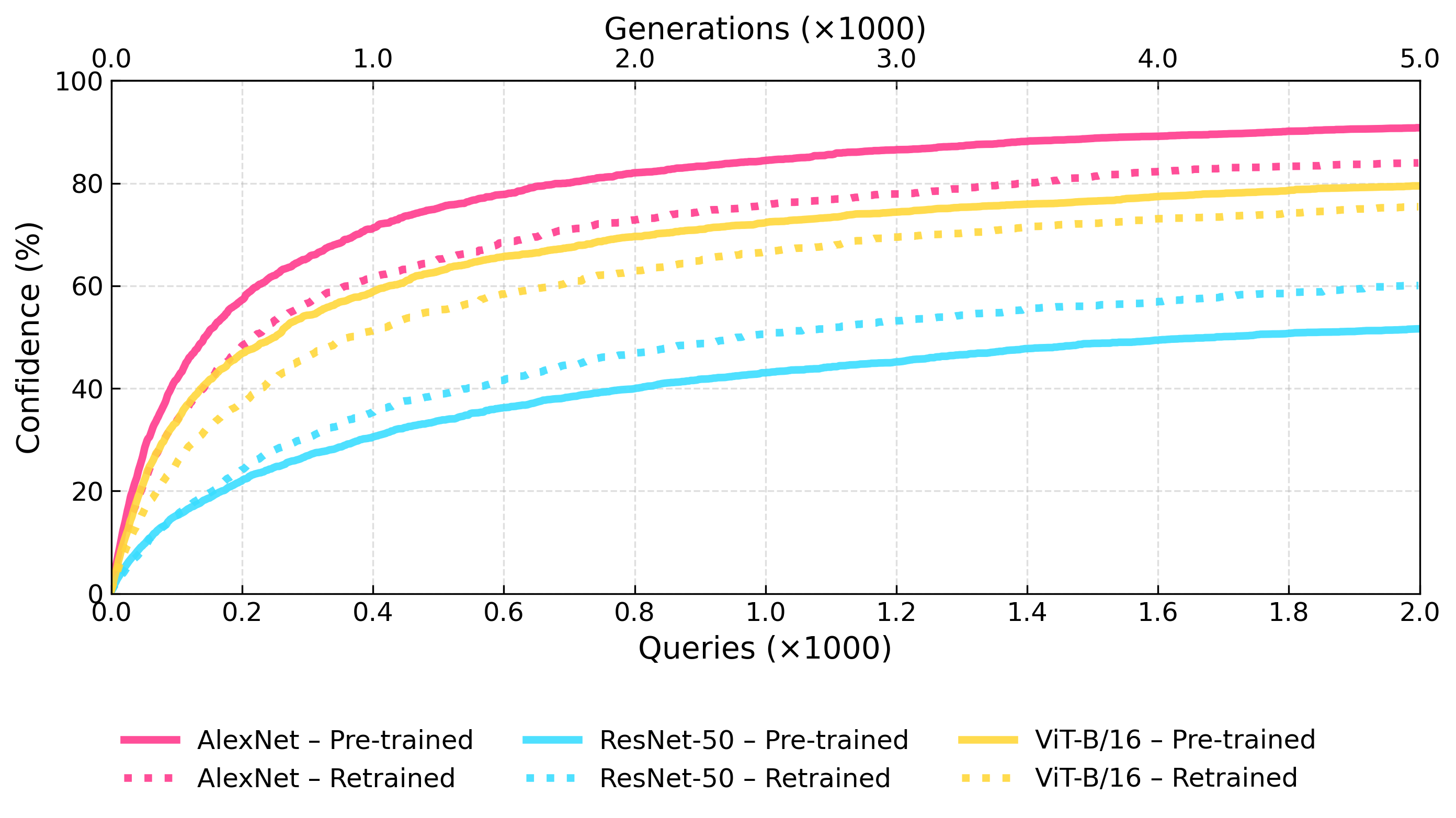}
    \caption{Fooling confidence vs Generations / Queries plot for \textbf{\cluneCPPNFool (Original)} re-implementation.}
    \label{fig:o_cppn_fool_plot}
    \vspace{-2mm}
\end{figure}

\begin{figure*}[ht]
    \centering
    \includegraphics[width=\textwidth]{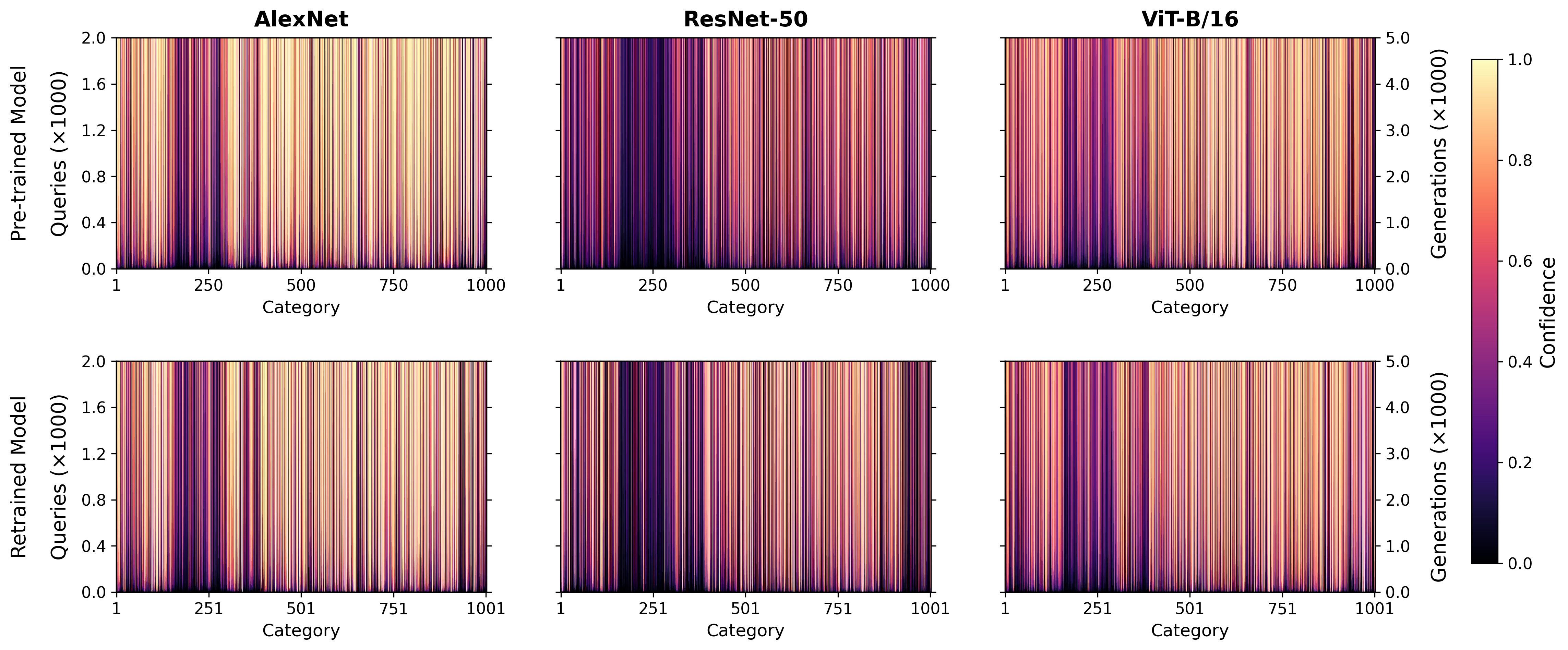}
    \caption{Fooling confidence heatmap across all \imageNetOneK classes for \textbf{\cluneCPPNFool (Original)} re-implementation.}
    \label{fig:o_cppn_fool_heatmap}
    \vspace{-2mm}
\end{figure*}

\begin{figure}[ht]
    \centering
    \includegraphics[width=\columnwidth]{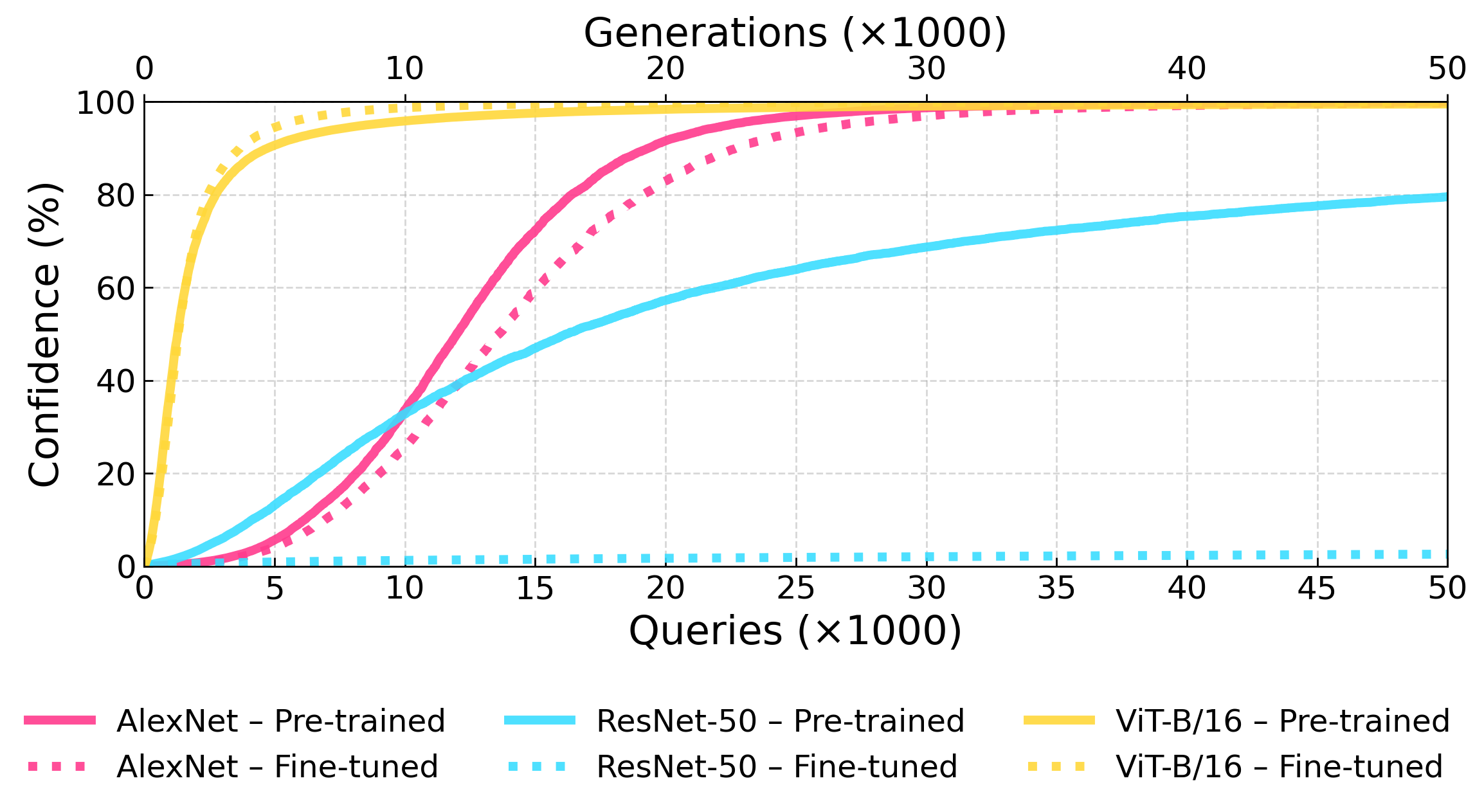}
    \caption{Fooling confidence vs Generations / Queries plot for \textbf{\cluneRawFool (Extended)} re-implementation.}
    \label{fig:e_raw_fool_plot}
    \vspace{-2mm}
\end{figure}

\begin{figure*}[t]
    \centering
    \includegraphics[width=\textwidth]{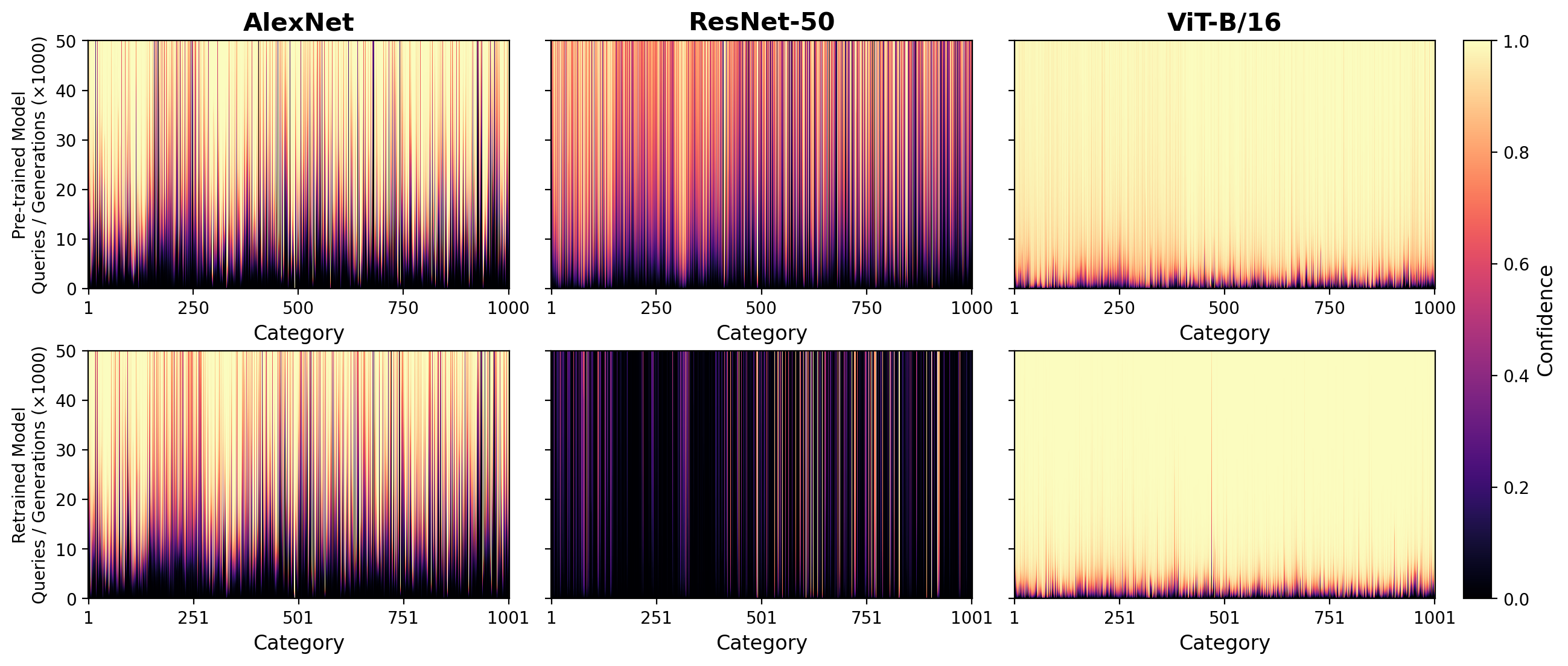}
    \caption{Fooling confidence heatmap across all \imageNetOneK classes for \textbf{\cluneRawFool (Extended)} re-implementation.}
    \label{fig:e_raw_fool_heatmap}
    \vspace{-2mm}
\end{figure*}

\begin{figure}[t]
    \centering
    \includegraphics[width=.95\columnwidth]{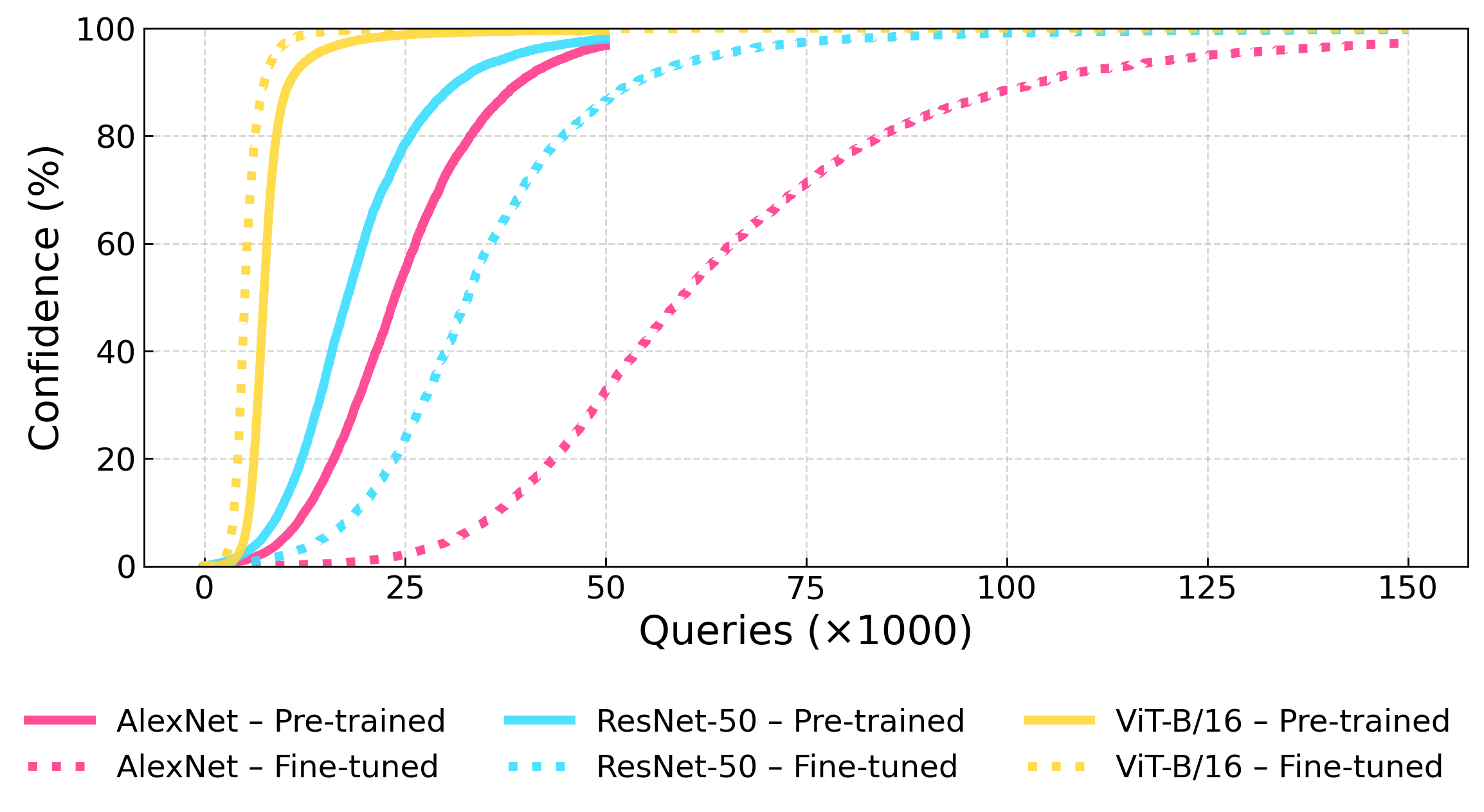}
    \caption{Fooling confidence vs Queries plot for \textbf{\methodName}}
    \label{fig:spoof_plot}
    \vspace{-2mm}
\end{figure}

\begin{figure*}[t]
    \centering
    \includegraphics[width=.95\textwidth]{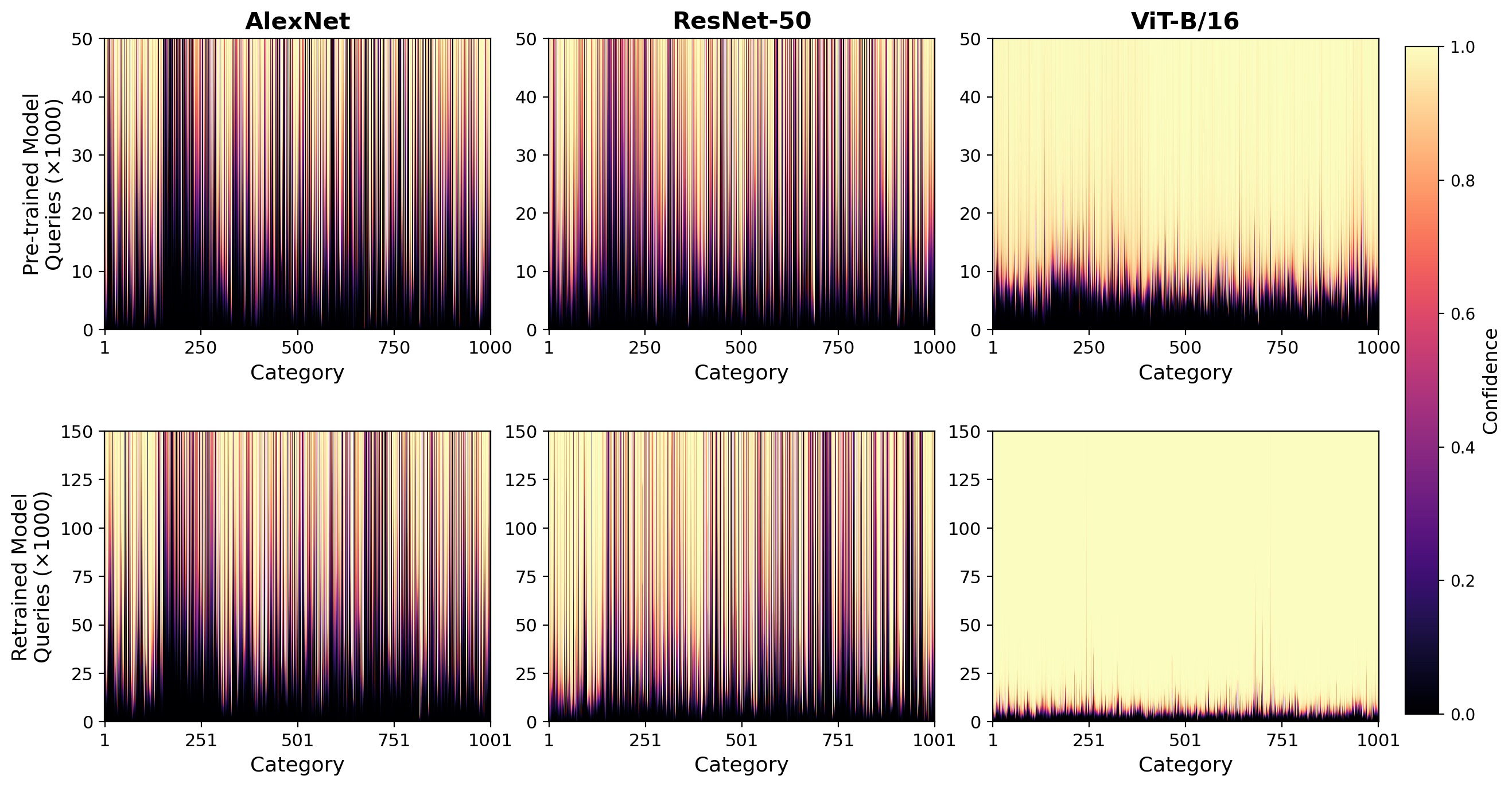}
    \caption{Fooling confidence heatmap across all \imageNetOneK classes for \textbf{\methodName}.}
    \label{fig:spoof_heatmap}
    \vspace{-2mm}
\end{figure*}

\section{Problem Setup}    
    Let $f: \mathbb{R}^{3 \times H \times W} \rightarrow \mathbb{R}^{N}$ be a pretrained image classifier and $y_t\!\in\!\{1,\dots,N\}$ a target class. The objective is to synthesize an input $\mathbf{x}$ from an empty canvas such that the classifier assigns high confidence to $y_t$, i.e., to maximize $f_{y_t}(\mathbf{x})$, using only black-box access to the classifier’s output probabilities, with minimal pixel changes to the initial image, i.e., the empty canvas.

    \begin{algorithm}[ht]
    \caption{\methodName}
    \label{alg:spoof}
    \begin{algorithmic}[1]
    \STATE \textbf{Input:} Classifier $f$, target class $c$, query budget $T$
    \STATE \textbf{Initialize:} Image $x \leftarrow 0$ (blank image)
    \STATE \phantom{\textbf{Initialize:}} Pixel Change $PC \leftarrow 0$ 
    \FOR{$t = 1$ to $T$}
        \STATE $x' \leftarrow x$
        \STATE Randomly sample pixel $(i,j)$ and channel $k$
        \STATE Sample value $v \sim \mathcal{U}(0,1)$
        \STATE $x'_{k,i,j} \leftarrow v$
        \IF{$f(x')_c > f(x)_c$}
            \STATE $x \leftarrow x'$ 
            \STATE $PC \leftarrow PC+1$
        \ENDIF
    \ENDFOR
    \STATE \textbf{Return:} Image $x$, Pixel Changes $PC$
    \end{algorithmic}
    \end{algorithm}

\begin{table*}[ht]
\centering
\small
\setlength{\tabcolsep}{6pt}
\renewcommand{\arraystretch}{1.0}
\caption{Fooling performance on retrained classifiers for the three fooling attacks. We chose to retrain \cluneCPPNFool (original) because of the long runtime required for data collection in the extended run, and because the original setting lead to satisfactory fooling performance. For \cluneRawFool, we selected the extended version for retraining because the original setting lead to very poor fooling performance.}
\begin{tabular}{l l c c c c}
\toprule
\textbf{Classifier} & \textbf{Fooling Attack} &
\textbf{Runtime (hours)} &
\textbf{Confidence (\%)} &
\textbf{Fooling-ASR (\%)} &
\textbf{\pcr (\%)} \\
\midrule

\multirow{3}{*}{AlexNet (Retrained)}
 & \cluneCPPNFool (Original)   & $2.87$ & $83.97$ & $90.91$  & $100$ \\
 & \cluneRawFool (Extended)    & $1.92$ & $99.72$ & $98.96$  & $100$ \\
 & \textbf{SPOOF (ours)}  & $\mathbf{0.50}$ & $\mathbf{97.24}$ & $\mathbf{92.57}$ & $\mathbf{3.09}$ \\
\midrule

\multirow{3}{*}{ResNet-50 (Retrained)}
 & \cluneCPPNFool (Original)   & $2.61$ & $60.05$ & $82.34$  & $100$ \\
 & \cluneRawFool (Extended)    & $2.97$ & $2.60$  & $35.40$  & $100$ \\
 & \textbf{SPOOF (ours)}  & $\mathbf{1.73}$ & $\mathbf{99.72}$ & $\mathbf{94.15}$ & $\mathbf{1.74}$ \\
\midrule

\multirow{3}{*}{ViT-B/16 (Retrained)}
 & \cluneCPPNFool (Original)   & $\mathbf{2.89}$ & $75.42$ & $93.23$  & $100$ \\
 & \cluneRawFool (Extended)    & $6.32$ & $99.95$ & $100.00$ & $100$ \\
 & \textbf{SPOOF (ours)}  & $4.76$ & $\mathbf{99.99}$ & $\mathbf{100.00}$ & $\mathbf{2.95}$ \\

\bottomrule
\end{tabular}
\label{tab:spoof_retrained}
\end{table*}

\begin{figure*}[t]
    \centering
    \includegraphics[width=0.95\textwidth]{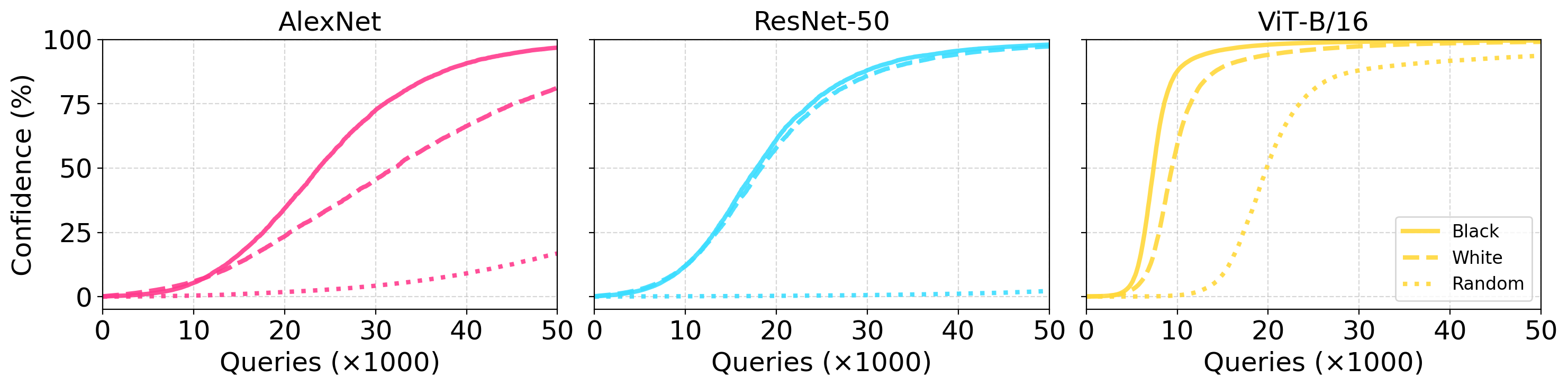}
    \caption{\textbf{\methodName Ablation}: Fooling confidence vs Generations plot for \methodName for three kinds of canvas initialization. We see the best fooling performance when starting from a black canvas, and the least fooling when starting with a random pixel canvas.}
    \label{fig:spoof_ablation}
    \vspace{-2mm}
\end{figure*}

\section{\methodName: Proposed Fooling Attack }
    \noindent
    \methodName builds on minimal-query adversarial attacks by evolving a single image sample per target. It reframes the task of fooling classifiers with unrecognizable inputs as a sparse, greedy search over pixel space, implemented through a lightweight single-sample hill-climbing procedure. Despite its simplicity—which contributes directly to both its efficiency and ease of deployment—\methodName proves remarkably effective. We describe the algorithm and its batched implementation below.
    
    \noIndentHeading{Initialization.}
    \methodName initializes a synthetic input as a uniform black canvas (\( \mathbf{x}\!=\!0 \)).
    
    \noIndentHeading{Greedy Pixel Updates.}
    At each iteration (``query-step''), \methodName selects a random pixel location $(x, y)$ and channel $c\!\in\!\{R,G,B\}$. 
    \methodName samples a random value $v$ from a uniform distribution $\mathcal{U}(0,1)$ and temporarily applies it to the selected pixel. 
    Then, it passes the updated image through the classifier to compute confidence scores $f_{y_t}(\mathbf{x})$. 
    If the modification improves the confidence, \methodName retains them. Despite its simplicity, this greedy strategy proves surprisingly effective, leveraging the smoothness in DNN decision surfaces. The updates continue for a default fixed query budget of $T$ steps. We describe this algorithm in \cref{alg:spoof}.
    
    \noIndentHeading{Batching.}
    For batch evaluation, \methodName initializes \( N \) uniform black canvases (\( \mathbf{x}\!=\!0 \)) to serve as initial synthetic inputs, with each one assigned to a distinct target class. Here, \( N \) corresponds to the total number of classes in the dataset (e.g., \( N{=}1{,}000 \) for \imageNetOneK and \( N{=}10 \) for \mnist datasets). These inputs can be viewed as \( N \) independent bins, evolved in parallel with updates applied independently but evaluated together in batch at each query step to leverage computational efficiency of GPUs. Since each image is updated independently in every step, the total number of queries per image is $T$, and the total number of queries overall is $N \times T$.

\section{Experiment Setup}
    \noIndentHeading{Datasets and Splits.}    
        Our experiments use the following datasets and splits:
        \begin{itemize}
            \item \imageNetOneK \val: This dataset contains $224\!\times\!224$ images belonging to $1K$ natural image categories. 
            The \val split contains about $1.3M$ and $50k$ images for training and testing, respectively.
            
            \item \mnist \val: This dataset contains $28\!\times\!28$ images of $10$ digits. 
            The \val split contains $50k$ and $10k$ images for training and testing, respectively.
        \end{itemize}
    
    \noIndentHeading{Classifiers.}
        We use the following classifiers for the above datasets:
        \begin{itemize}
            \item \imageNetOneK: \alexNet \cite{krizhevsky2012imagenet}, \resNetFifty \cite{he2016deep}, and \vitB \cite{dosovitskiy2020image} were chosen as these classifiers encompasses both convolution and transformer-based architectures. We use their \textit{torchvision} pretrained checkpoints with their weights fixed during optimization.

            \item \mnist: We use LeNet \cite{lecun1998gradient} generating fooling images by \methodName, given the same was reported for \cluneCPPNFool and \cluneRawFool by \cite{nguyen2015deep}.
        \end{itemize}

    \noIndentHeading{Hyperparameters.}
        Unless stated otherwise, all \methodName experiments on \imageNetOneK use a query budget of $T{=}50{,}000$, batch size $N$ equal to the number of dataset classes. We initialize all fooling images as black canvases and evolve without any $\ell_p$ constraint.
    
    \noIndentHeading{Baseline.}
        We compare \methodName against re-implementation of both fooling attacks introduced by \cite{nguyen2015deep} with the original and extended parameter setting. The first, \cluneCPPNFool, uses compositional pattern-producing networks (CPPNs, an indirect encoding), while the second, \cluneRawFool, operates directly in pixel space (direct encoding). The CPPN-based indirect encoding can evolve complex, globally regular images rich in naturalistic and man-made features~\cite{stanley2007compositional,secretan2008picbreeder,auerbach2012automated}.

    \noIndentHeading{Nguyen et al. (2015) Re-implementation.} 
        As the fooling images introduced in \cite{nguyen2015deep} were only generated on \imageNetOneK trained \alexNet, in our re-implementation, we first reproduced results on \alexNet and then used the same experiment settings to generate fooling images on state-of-the-art classifiers that came after \alexNet, i.e., \resNetFifty and \vitB. As the two fooling attacks introduced in \cite{nguyen2015deep} -- \cluneCPPNFool and \cluneRawFool -- were implemented for different parameter settings (see \cref{tab:parameters}), we extended the original experiment settings in \cite{nguyen2015deep} and ran experiments with a common query budget per class for both fooling attacks from \cite{nguyen2015deep} as well as using the same query budget with our proposed fooling attack,  \methodName. Thus, in the re-implementation, we conducted 4 independent experiments (using the 3 classifiers), that we call, 
        \begin{itemize}
            \item \cluneCPPNFool (Original) and \cluneCPPNFool (Extended)
            \item \cluneRawFool (Original) and \cluneRawFool (Extended)
        \end{itemize}

    \noIndentHeading{Retraining with Fooling Images.}
        To assess classifier robustness on retraining, each \imageNetOneK-pretrained classifier was fine-tuned using $12{,}000$ corresponding fooling images ($10$k train / $2$k val), appended as a new ($1001^{\text{st}}$) class following \cite{nguyen2015deep}. Fine-tuning was limited to the final classification layer, with all pretrained weights kept frozen. Baseline validation accuracy was recovered within ten epochs, and the epoch-10 classifiers were retained as the final retrained classifiers for re-evaluation.

    \section{Results}
    \subsection{\imageNetOneK Experiments}

    \noIndentHeading{\methodName: the most efficient, consistent, and simple fooling attack.} As reported in \cref{tab:spoof_results}, \methodName attains comparable or higher median confidence than \cluneCPPNFool and \cluneRawFool while changing $<3\%$ of pixels on average. \methodName requires considerably less wall time for all except a single attack setting, yet still reaches near-certain confidence across targets (see the per-target curves and plateaus in \cref{fig:spoof_plot}). Representative outputs show SPOOF produces unrecognizable, high-confidence images with only sparse pixel changes (see examples in \cref{fig:poster}). The efficiency gain follows directly from SPOOF’s greedy single-sample simple hill-climb: it avoids population management, crossover and mutation bookkeeping, and thus converts query budget into steady confidence gains rather than the more expensive optimizations needed for methods by \cite{nguyen2015deep}.


    \noIndentHeading{\vitB: the most vulnerable classifier to fooling attacks.}
    Across nearly all attack settings—except for CPPN-Fool, where ViT-B/16 remains comparably vulnerable to AlexNet—\vitB{} emerges as the easiest model to fool. Under \methodName, ViT-B/16 reaches a median target confidence of 95\% after only 13.7k queries and ultimately attains 99.6\% (\cref{fig:spoof_plot}, \cref{tab:spoof_results}), substantially faster than AlexNet or ResNet-50. Per-class heatmaps reinforce this trend: ViT exhibits broad, early confidence growth across categories, whereas convolutional models require far more queries or extended evolutionary runs (\cref{fig:e_raw_fool_heatmap}, \cref{fig:spoof_heatmap}). Despite strong in-distribution performance, ViT shows pronounced overconfidence on unrecognizable, OOD inputs—a vulnerability \methodName exploits most efficiently.

    

    \noIndentHeading{Fine-tuning offers limited resilience to fooling attacks.}
    Retraining the classifiers with an additional “fooling” class delays \methodName{}ing but does not prevent it; the method eventually regains high-confidence fooling across all models. Other attacks similarly remain effective after fine-tuning, with the sole exception of \resNetFifty{} under \cluneRawFool{} (\cref{fig:o_cppn_fool_plot}, \cref{fig:e_raw_fool_plot}, \cref{fig:spoof_plot}, \cref{tab:spoof_retrained}). These results suggest that despite good retraining validation performance the decision boundaries responsible for fooling are not easily reshaped through standard fine-tuning and that the backbone does not rely solely on stable low-level features when confronted with unrecognizable, non-semantic inputs.
    
    \noIndentHeading{\methodName{} ablation: white-canvas initialization still yields high-confidence fooling.}
    As shown in \cref{fig:spoof_ablation}, starting from a white canvas also produces high-confidence fooling, though with a modest delay compared to black initialization. Notably, under black initialization, \resNetFifty{} exhibits strong resistance even without retraining, whereas \vitB{} remains highly susceptible to \methodName. The relative vulnerability pattern closely mirrors that observed for \cluneRawFool{} (\cref{fig:e_raw_fool_plot}), consistent with both setups exploring effectively full pixel-space noise.

    \subsection{\mnist Experiments}
    For the \mnist-trained LeNet, \cluneCPPNFool and \cluneRawFool reached the reported near-perfect confidence of ($99.99\%$) \cite{nguyen2015deep}, whereas \methodName achieves comparable fooling confidence ($99.68\%$) with just $21.08\%$ pixel change to the blank canvas and about eightfold fewer queries. See supplementary material for more details.

    \section{Conclusion}
    We introduced \methodName{}, a minimalistic black-box fooling attack that exposes persistent overconfidence in modern image classifiers on out-of-distribution unrecognizable (fooling) images. Through extensive experiments across \alexNet{}, \resNetFifty{}, and \vitB{}, we showed that \methodName{} achieves comparable or higher fooling confidence than \cite{nguyen2015deep}, the only other existing fooling attack at \imageNetOneK scale, while being faster and altering a small fraction of pixels on average. Our analysis revealed that even state-of-the-art transformers like \vitB{} are highly susceptible, and that retraining with fooling-aware objectives offers no meaningful resilience. Qualitative and ablation studies further confirmed that neither structural priors nor initialization choices are necessary for inducing high-confidence fooling. These findings highlight a critical gap between recognition performance and OOD robustness—current visual models remain easily deceived by unrecognizable stimuli, emphasizing the urgent need for architectures and training paradigms that reason beyond confidence alone.

{
    \small
    \bibliographystyle{ieeenat_fullname}

}

\twocolumn[{%
\begin{center}
\Large{\bf Supplementary Material for\\
\methodName: Simple Pixel Operations for Out-of-Distribution Fooling}\\
\vspace{1cm}
\normalsize{Ankit Gupta, Christoph Adami, and Emily Dolson}
\end{center}
}]
\mbox{}
\setcounter{figure}{0}
\section*{Overview}
This supplementary document includes additional fooling examples on various ImageNet classes—extending the illustration of fooling images shown in the main paper at Fig. 1 (right)—along with mean-aggregated ImageNet-1K metrics across 1000 targets, ablations of \methodName{} under different initialization settings, and MNIST experiments. Collectively, these results provide broader context and further reinforce the conclusions of the main paper.

\section*{S1: \imageNetOneK Results}
\subsection*{S1.1: Additional Fooling Images Across Classes}
We provide additional fooling examples for a variety of ImageNet classes across all three fooling attacks and classifiers. These examples extend the comparison shown in Figure 1 (Right) of the main paper and highlight the qualitative differences between CPPN-Fool, Direct-Fool, and \methodName (see \cref{%
fig:fooling_489,%
fig:fooling_562,%
fig:fooling_646,%
fig:fooling_750,%
fig:fooling_772,%
fig:fooling_794,%
fig:fooling_858,%
fig:fooling_868,%
fig:fooling_885,%
fig:fooling_971})

\begin{figure}[ht]
    \centering
    \includegraphics[width=\linewidth]{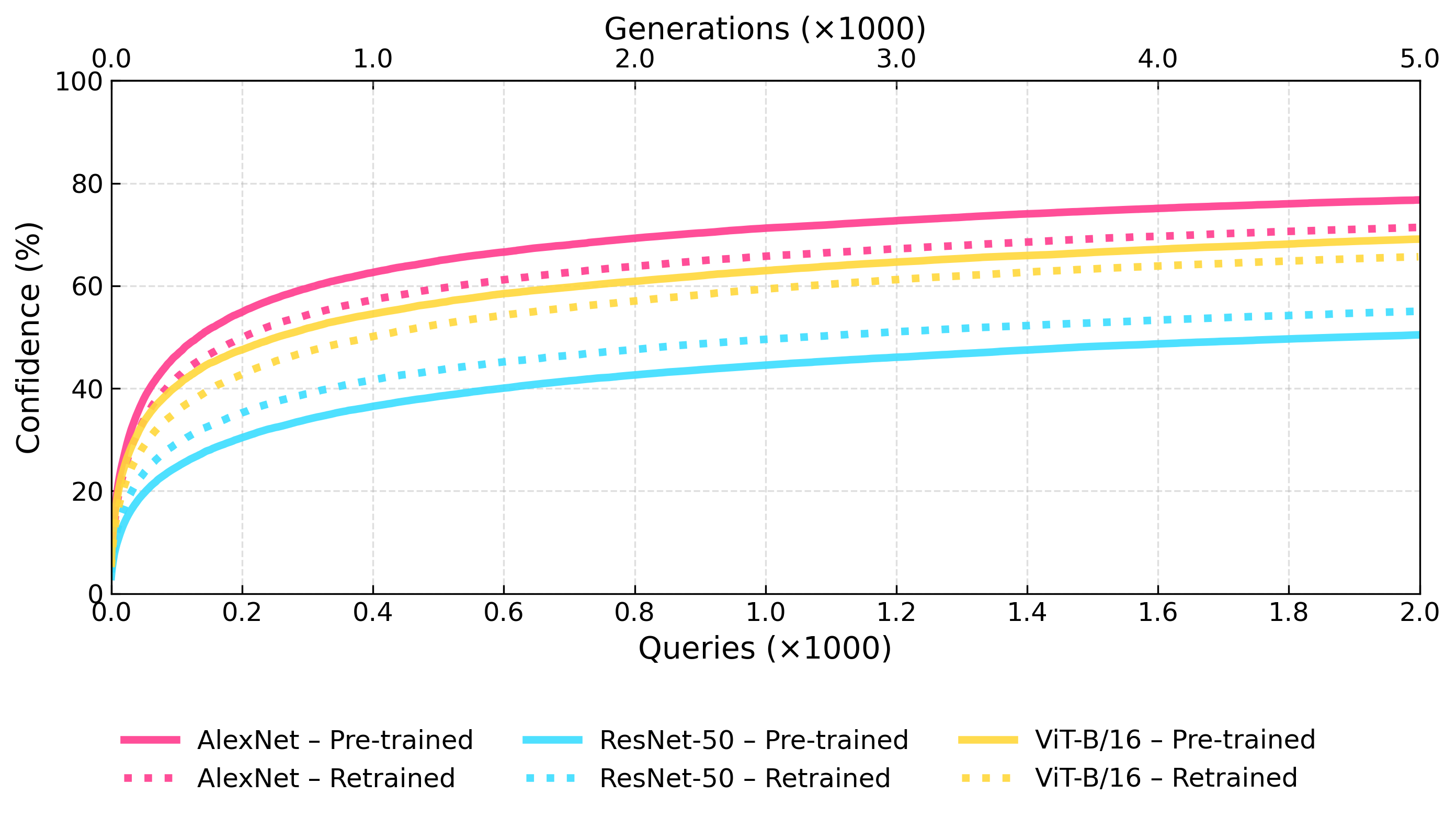}
    \caption{
        \textbf{Mean confidence for CPPN-Fool} across 1000 ImageNet target classes.
    }
    \label{fig:mean_cppn}
\end{figure}

\begin{figure}[ht]
    \centering
    \includegraphics[width=\linewidth]{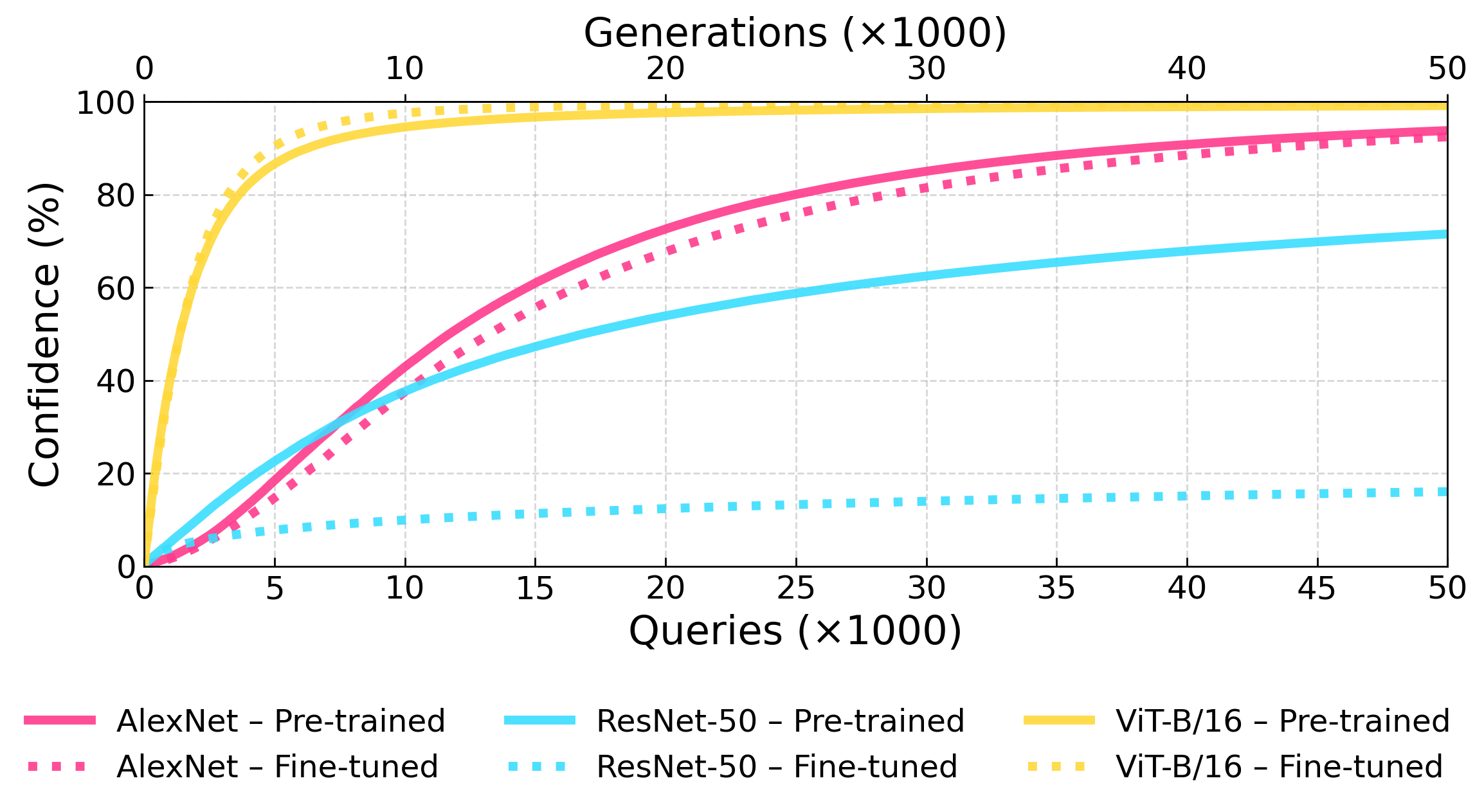}
    \caption{
        \textbf{Mean confidence for Direct-Fool} across 1000 ImageNet target classes.
    }
    \label{fig:mean_raw}
\end{figure}

\begin{figure}[ht]
    \centering
    \includegraphics[width=\linewidth]{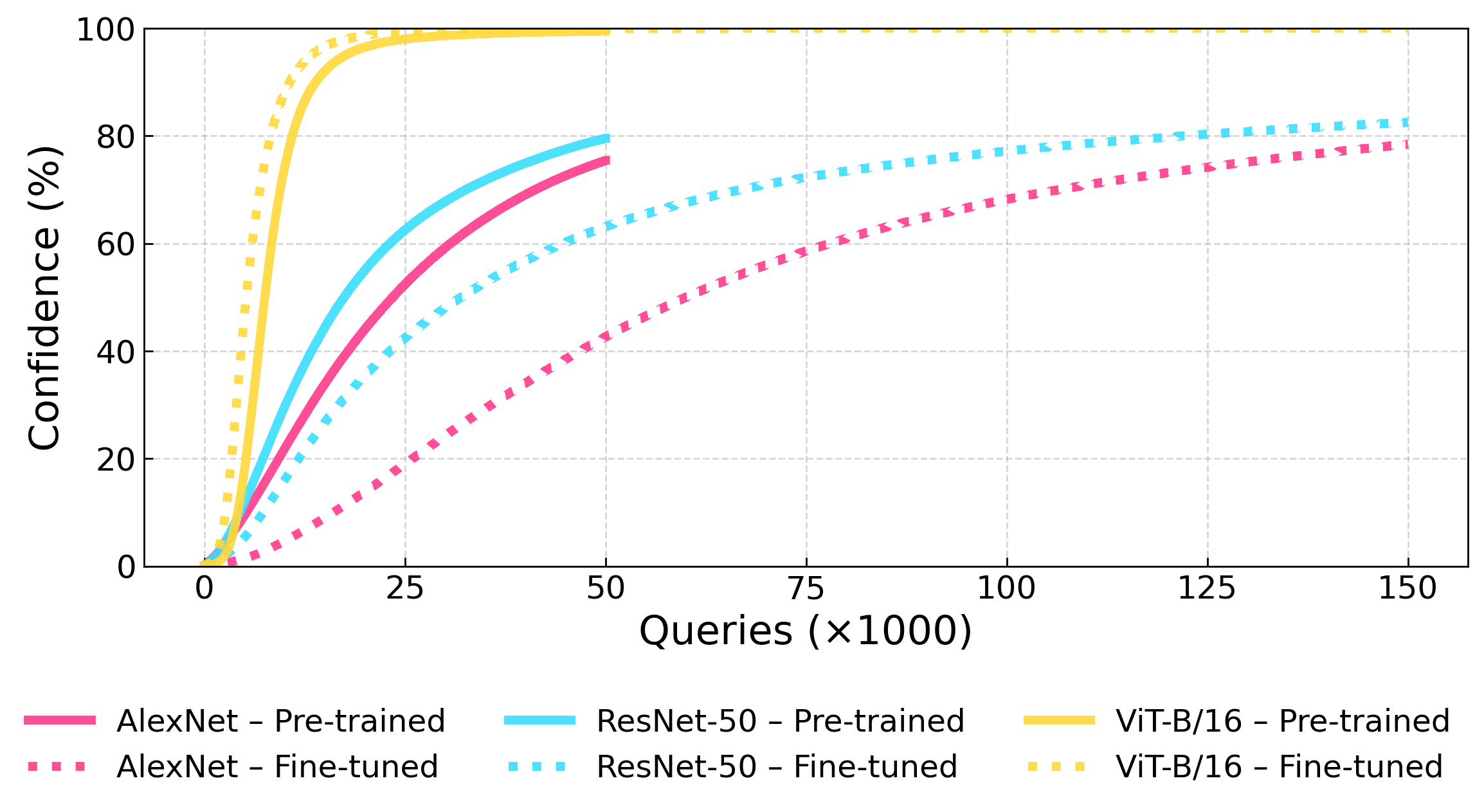}
    \caption{
        \textbf{Mean confidence for \methodName{}} across 1000 ImageNet target classes.
    }
    \label{fig:mean_spoof}
\end{figure}

\begin{table*}[ht]
\centering
\small
\setlength{\tabcolsep}{8pt}
\renewcommand{\arraystretch}{1.2}
\caption{Mean fooling confidence and Pixel Change Ratio (PCR) across 1000 ImageNet-1K target classes, evaluated on pretrained ImageNet classifiers.}
\begin{tabular}{l l c c}
\toprule
\textbf{Classifier} & \textbf{Fooling Attack} & \textbf{Confidence (\%)} & \textbf{PCR (\%)} \\
\midrule

\multirow{5}{*}{AlexNet}
 & CPPN-Fool (Original)  & 76.74 & 100 \\
 & CPPN-Fool (Extended)  & 91.43 & 100 \\
 & Direct-Fool (Original) & 37.83 & 100 \\
 & Direct-Fool (Extended) & \textbf{93.76} & 100 \\
 & \textbf{SPOOF (ours)} & 75.42 & \textbf{2.31} \\
\midrule

\multirow{5}{*}{ResNet-50}
 & CPPN-Fool (Original)  & 50.44 & 100 \\
 & CPPN-Fool (Extended)  & 72.77 & 100 \\
 & Direct-Fool (Original) & 35.02 & 100 \\
 & Direct-Fool (Extended) & 71.52 & 100 \\
 & \textbf{SPOOF (ours)} & \textbf{79.53} & \textbf{1.23} \\
\midrule

\multirow{5}{*}{ViT-B/16}
 & CPPN-Fool (Original)  & 69.10 & 100 \\
 & CPPN-Fool (Extended)  & 87.23 & 100  \\
 & Direct-Fool (Original) & 93.49 & 100 \\
 & Direct-Fool (Extended) & 99.13 & 100 \\
 & \textbf{SPOOF (ours)} & \textbf{99.41} & \textbf{3.63} \\
\bottomrule
\end{tabular}
\label{tab:mean_results}
\end{table*}

\begin{table*}[ht]
\centering
\small
\setlength{\tabcolsep}{8pt}
\renewcommand{\arraystretch}{1.2}
\caption{Mean fooling confidence and Pixel Change Ratio (PCR) across 1000 ImageNet-1K target classes for \textbf{retrained} classifiers (using an additional ``fooling'' class).}
\begin{tabular}{l l c c}
\toprule
\textbf{Classifier (Retrained)} & \textbf{Fooling Attack} & \textbf{Confidence (\%)} & \textbf{PCR (\%)} \\
\midrule

\multirow{3}{*}{AlexNet}
 & CPPN-Fool (Original)      & 71.39  & 100 \\
 & Direct-Fool (Extended)    & \textbf{92.43}  & 100 \\
 & \textbf{SPOOF (ours)}     & {78.42} & \textbf{3.03} \\
\midrule

\multirow{3}{*}{ResNet-50}
 & CPPN-Fool (Original)      & 55.04  & 100 \\
 & Direct-Fool (Extended)    & 16.06  & 100 \\
 & \textbf{SPOOF (ours)}     & \textbf{82.47} & \textbf{1.73} \\
\midrule

\multirow{3}{*}{ViT-B/16}
 & CPPN-Fool (Original)      & 65.67  & 100 \\
 & Direct-Fool (Extended)    & 99.89  & 100 \\
 & \textbf{SPOOF (ours)}     & \textbf{99.99} & \textbf{3.02} \\
\bottomrule
\end{tabular}
\label{tab:mean_retrained_results}
\end{table*}

\begin{figure*}[t]
\centering
\includegraphics[width=\textwidth]{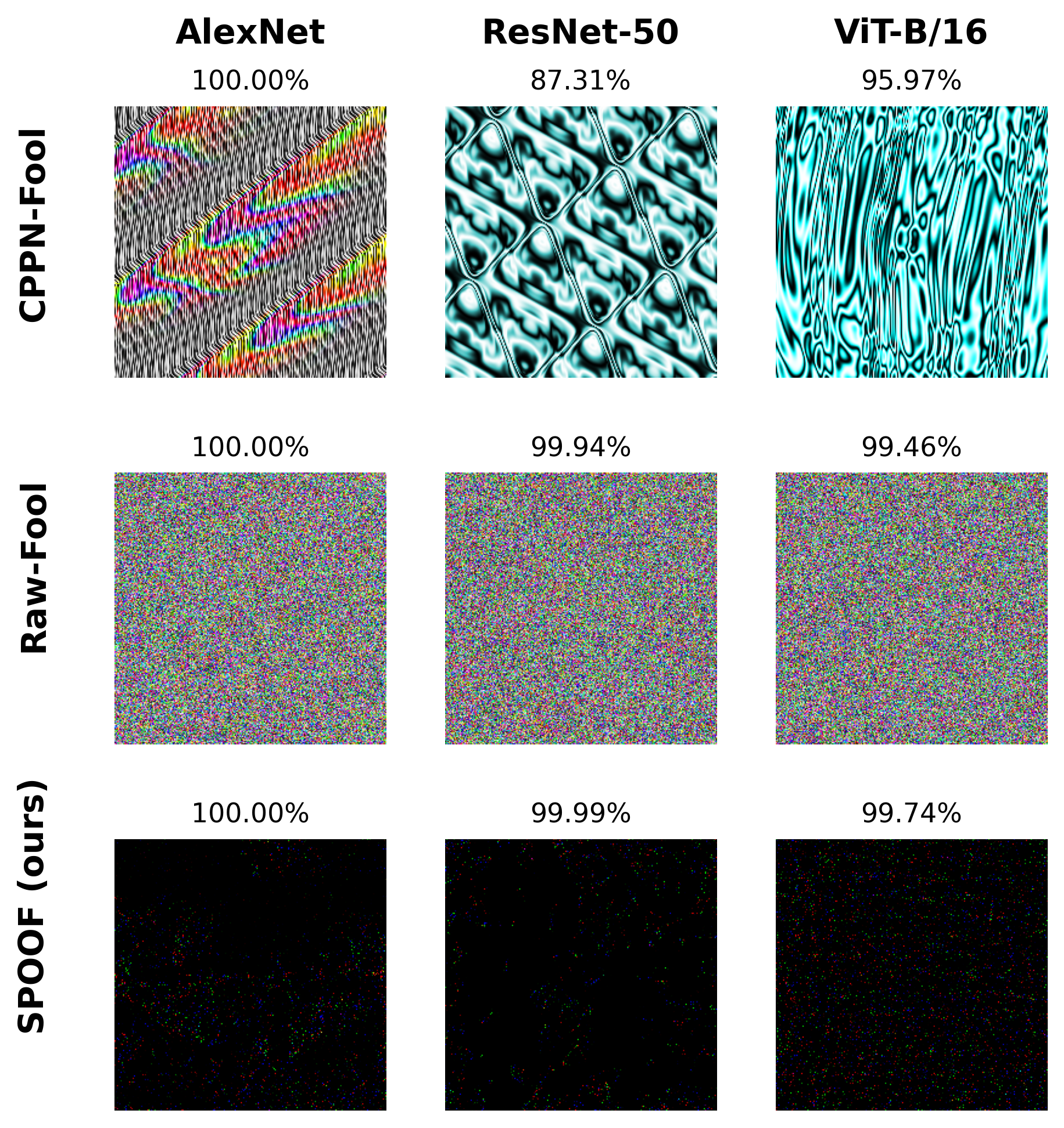}
\caption{Fooling images generated for \textbf{Class 489 — Chainlink Fence}.}
\label{fig:fooling_489}
\end{figure*}

\begin{figure*}[t]
\centering
\includegraphics[width=\textwidth]{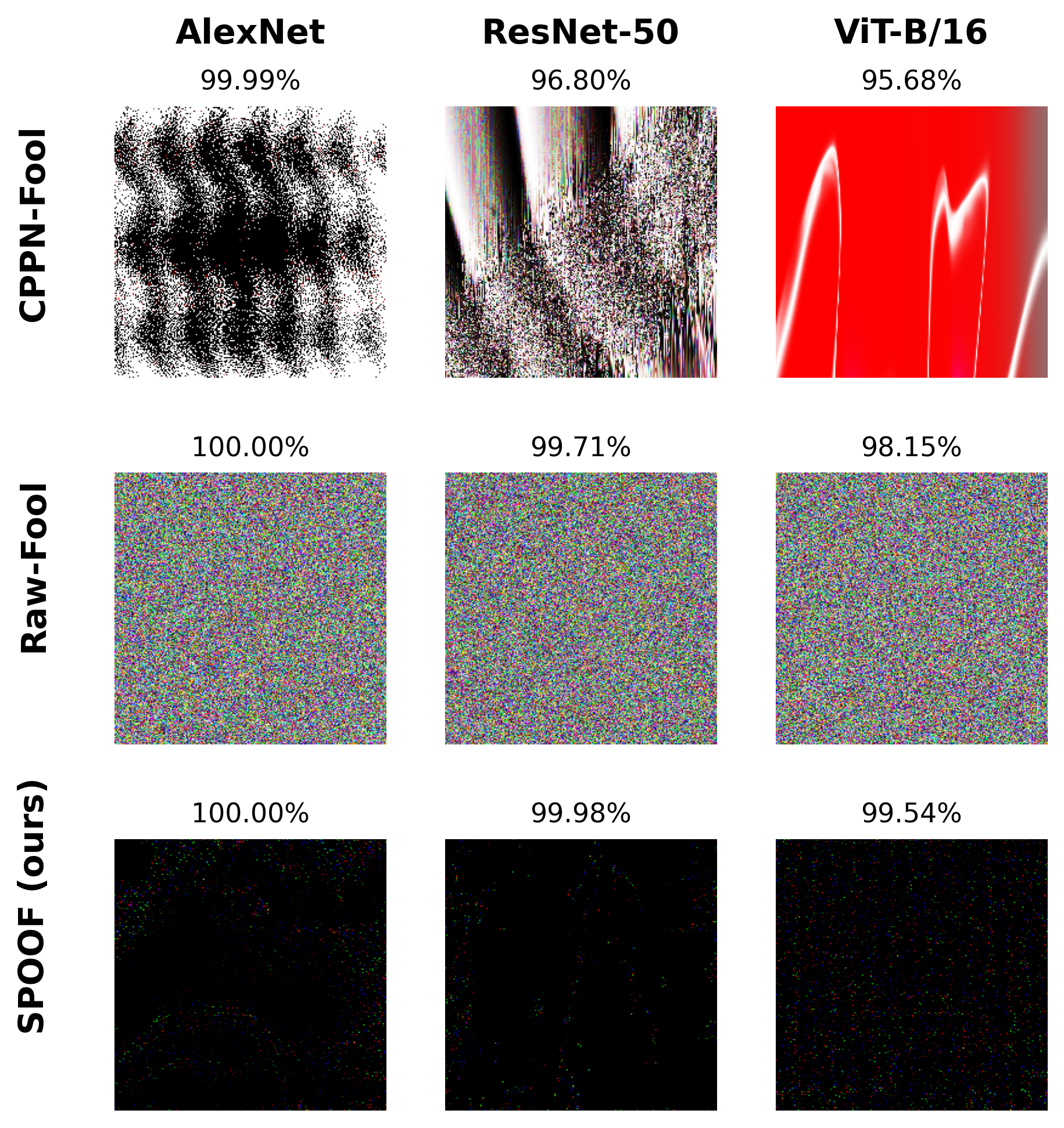}
\caption{Fooling images generated for \textbf{Class 562 — Fountain}.}
\label{fig:fooling_562}
\end{figure*}

\begin{figure*}[t]
\centering
\includegraphics[width=\textwidth]{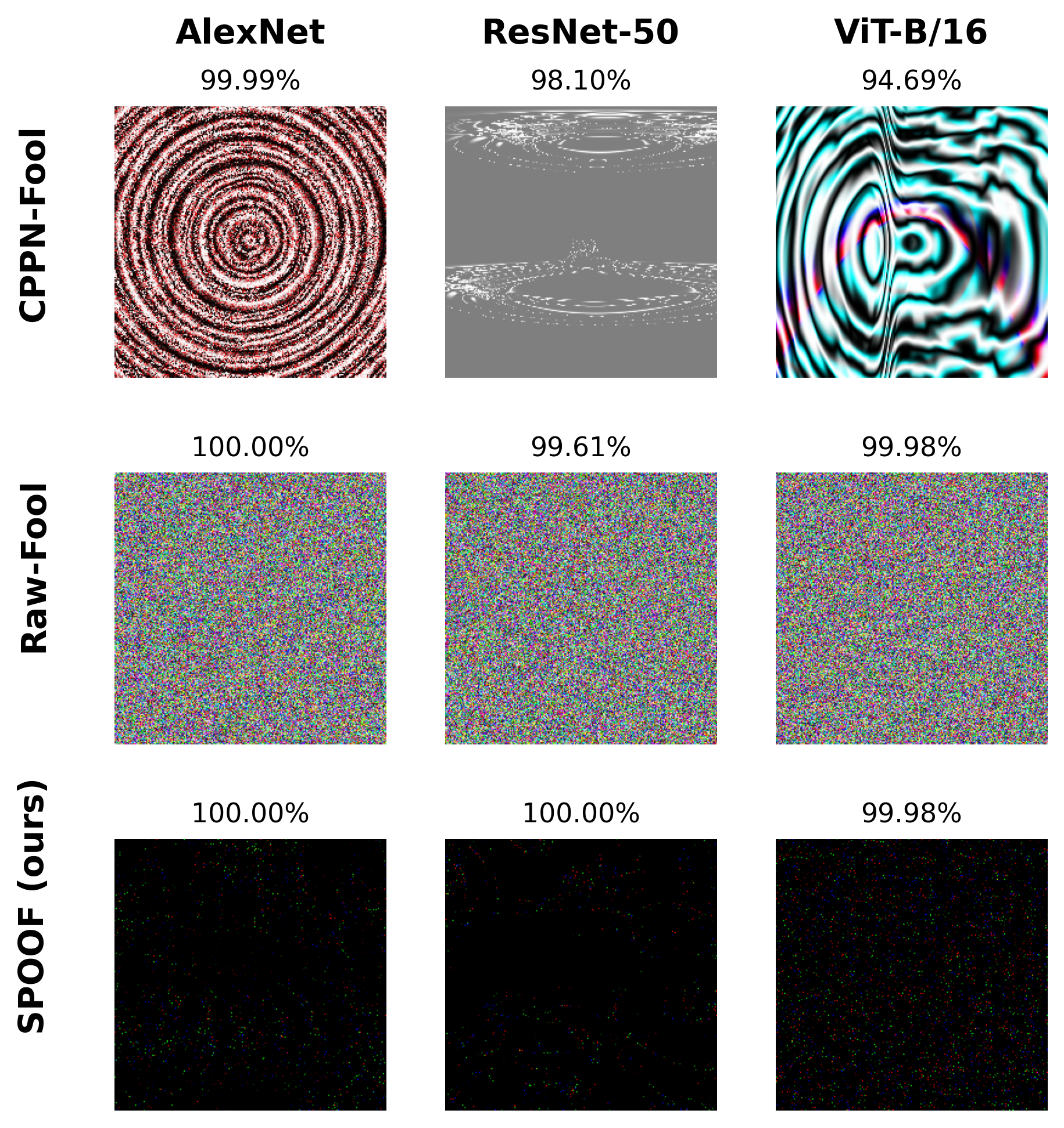}
\caption{Fooling images generated for \textbf{Class 646 — Maze}.}
\label{fig:fooling_646}
\end{figure*}

\begin{figure*}[t]
\centering
\includegraphics[width=\textwidth]{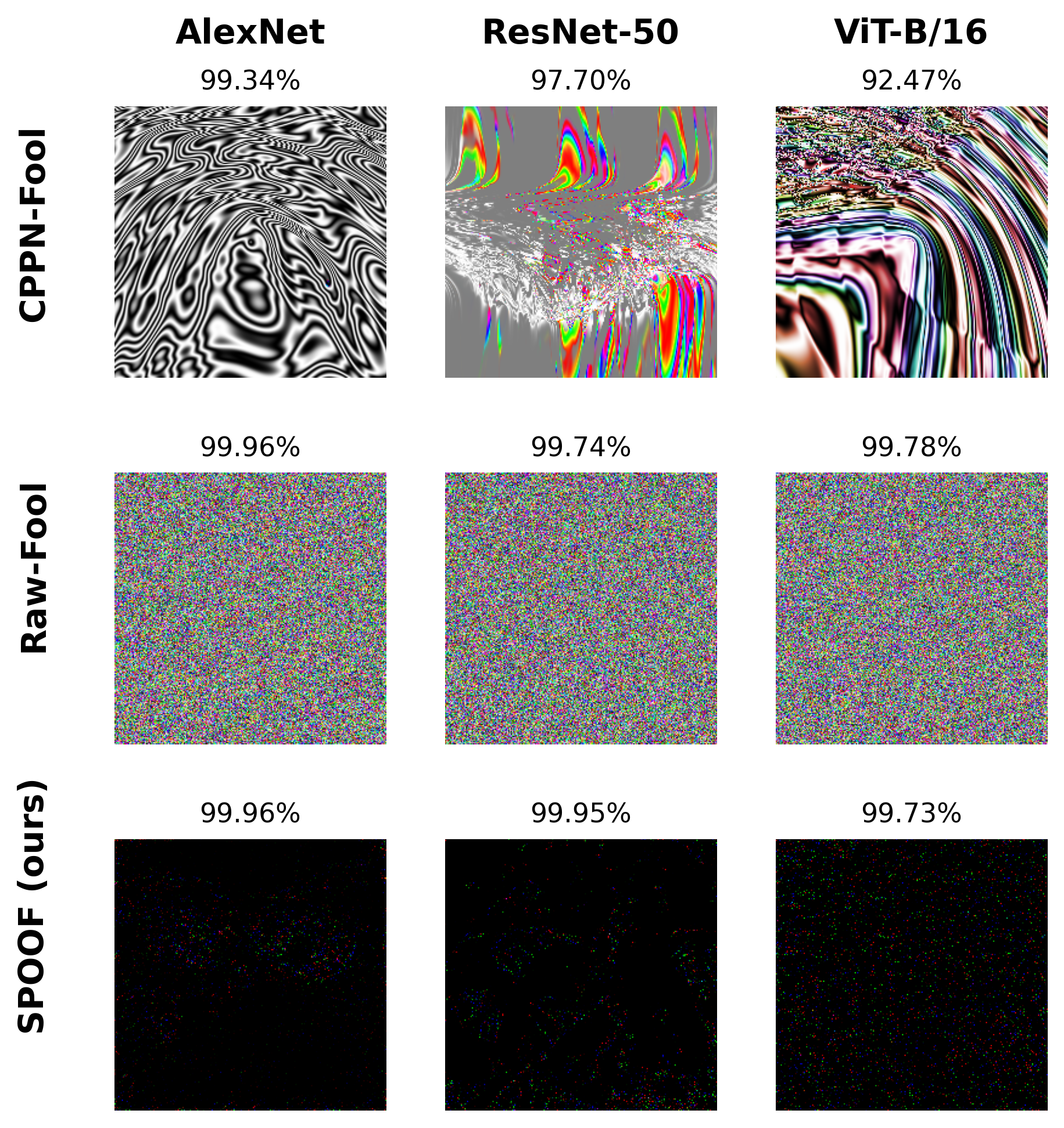}
\caption{Fooling images generated for \textbf{Class 750 — Quilt}.}
\label{fig:fooling_750}
\end{figure*}

\begin{figure*}[t]
\centering
\includegraphics[width=\textwidth]{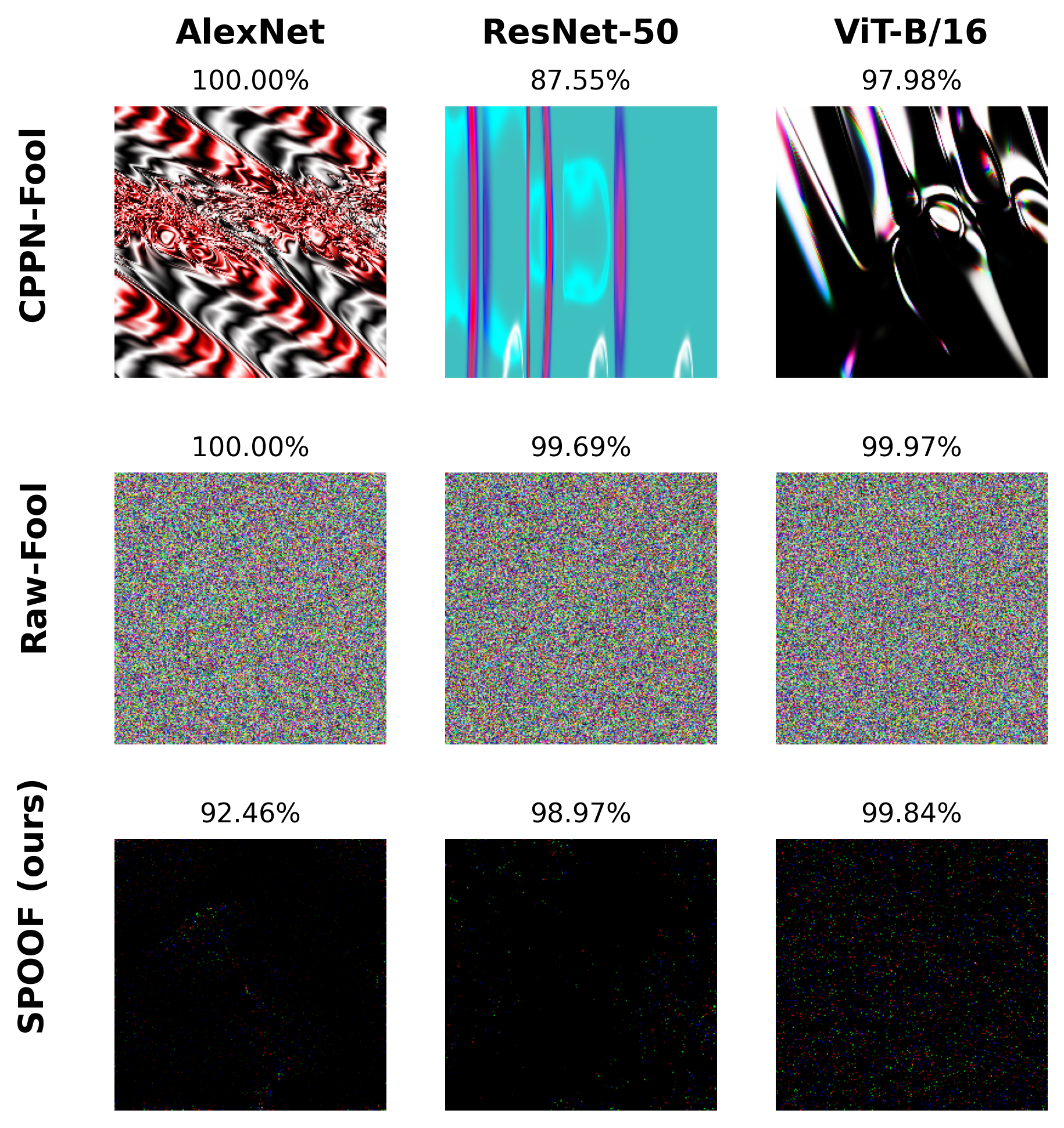}
\caption{Fooling images generated for \textbf{Class 772 — Safety Pin}.}
\label{fig:fooling_772}
\end{figure*}

\begin{figure*}[t]
\centering
\includegraphics[width=\textwidth]{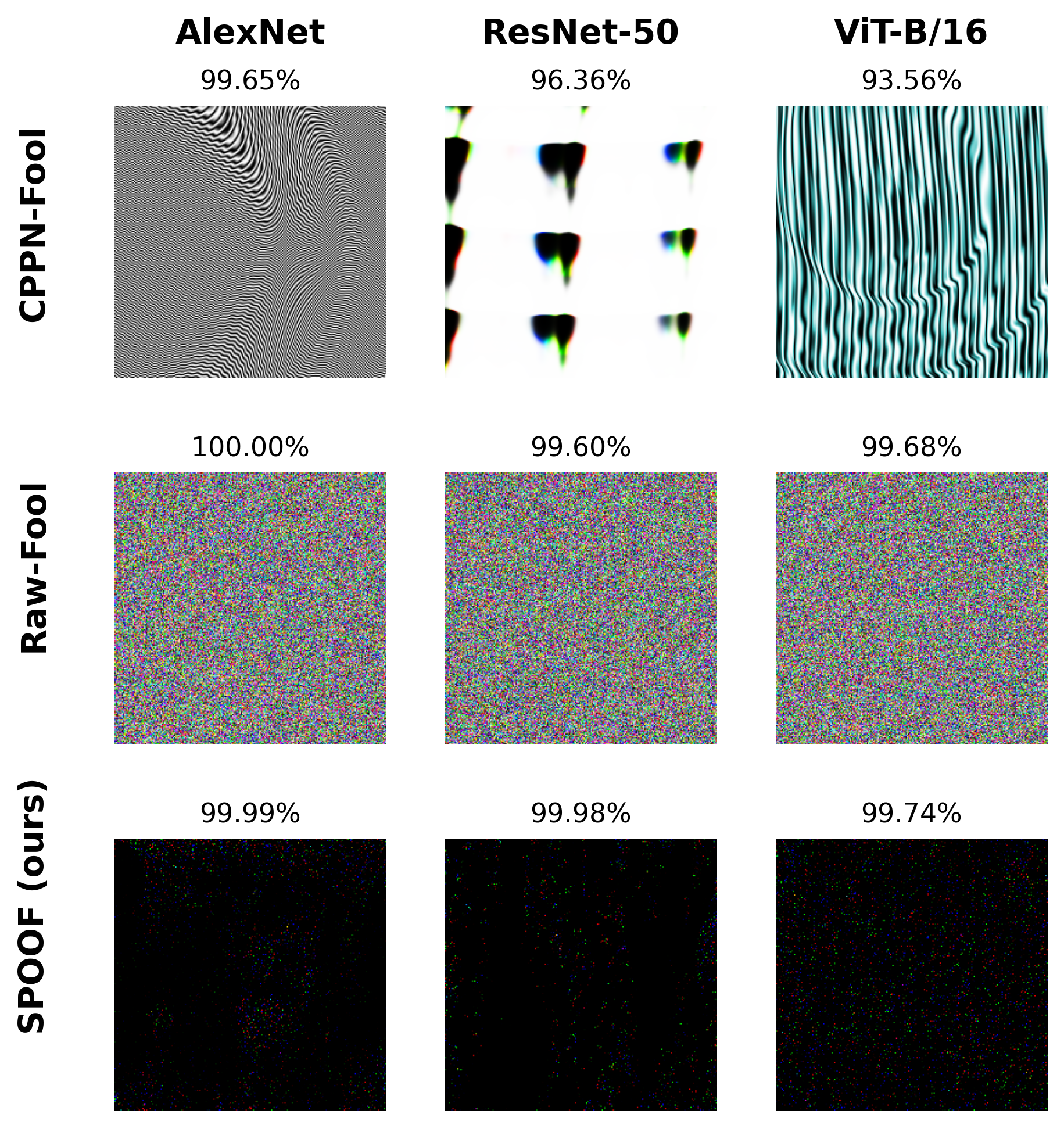}
\caption{Fooling images generated for \textbf{Class 794 — Shower Curtain}.}
\label{fig:fooling_794}
\end{figure*}

\begin{figure*}[t]
\centering
\includegraphics[width=\textwidth]{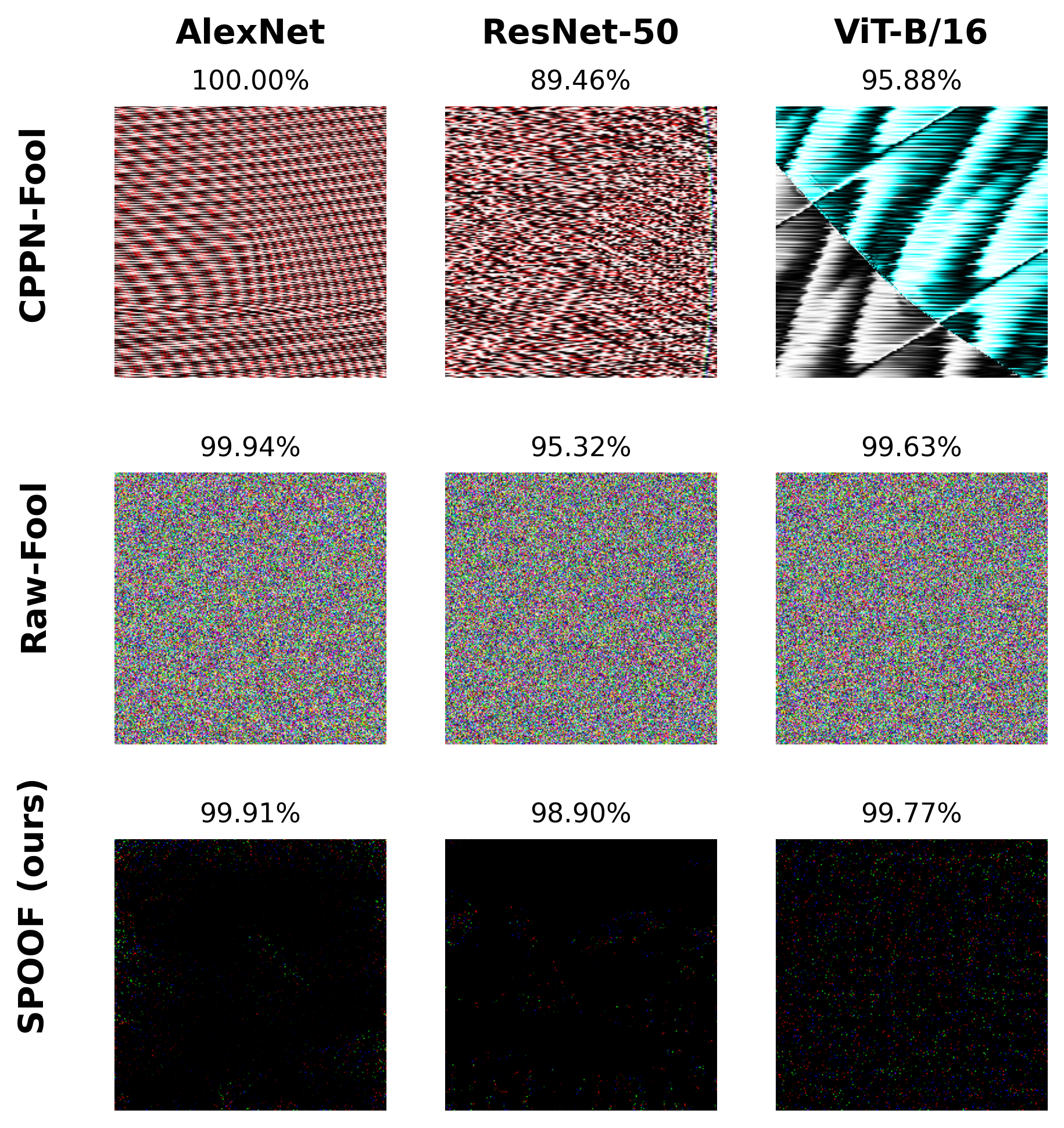}
\caption{Fooling images generated for \textbf{Class 858 — Tile Roof}.}
\label{fig:fooling_858}
\end{figure*}

\begin{figure*}[t]
\centering
\includegraphics[width=\textwidth]{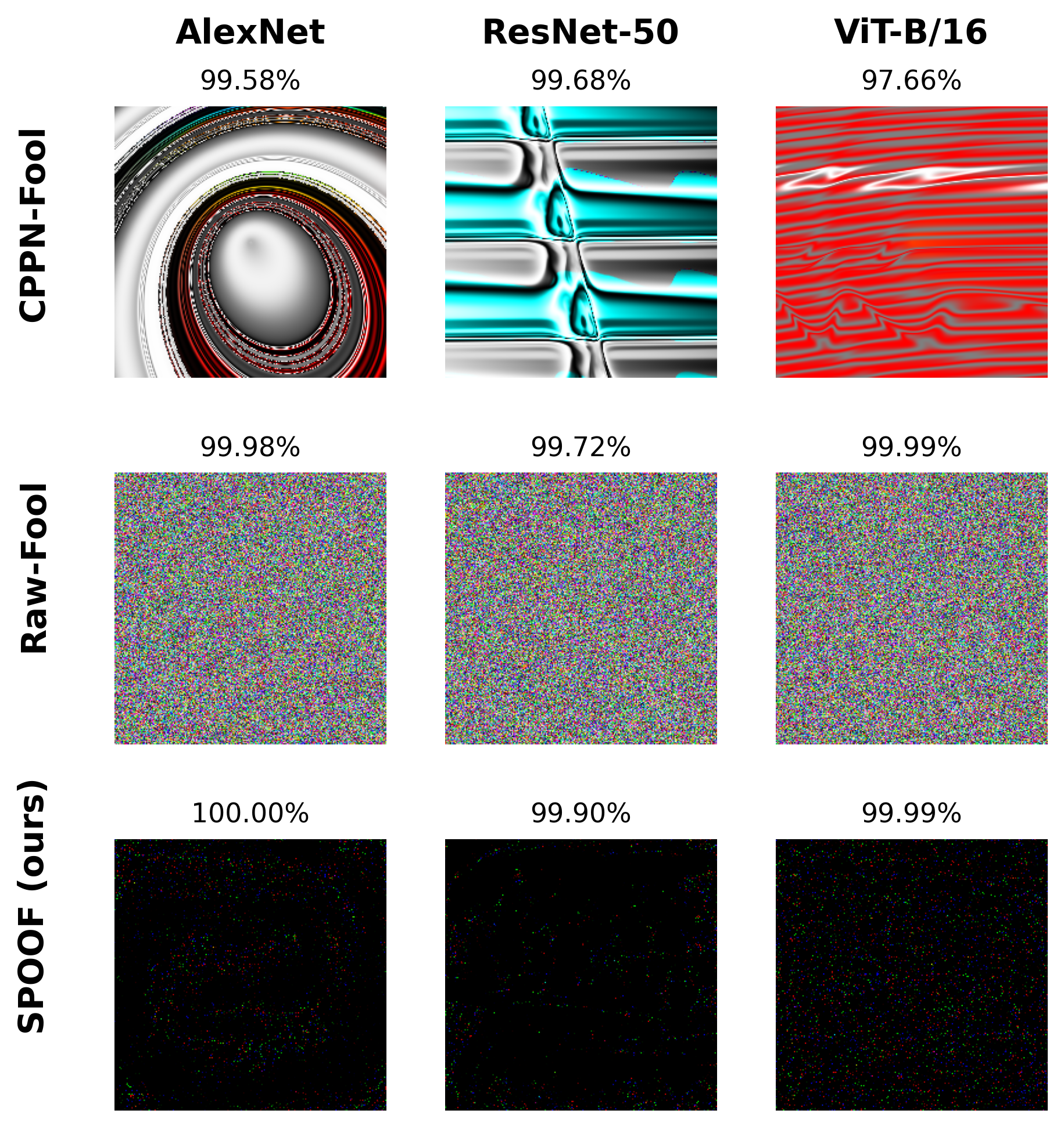}
\caption{Fooling images generated for \textbf{Class 868 — Tray}.}
\label{fig:fooling_868}
\end{figure*}

\begin{figure*}[t]
\centering
\includegraphics[width=\textwidth]{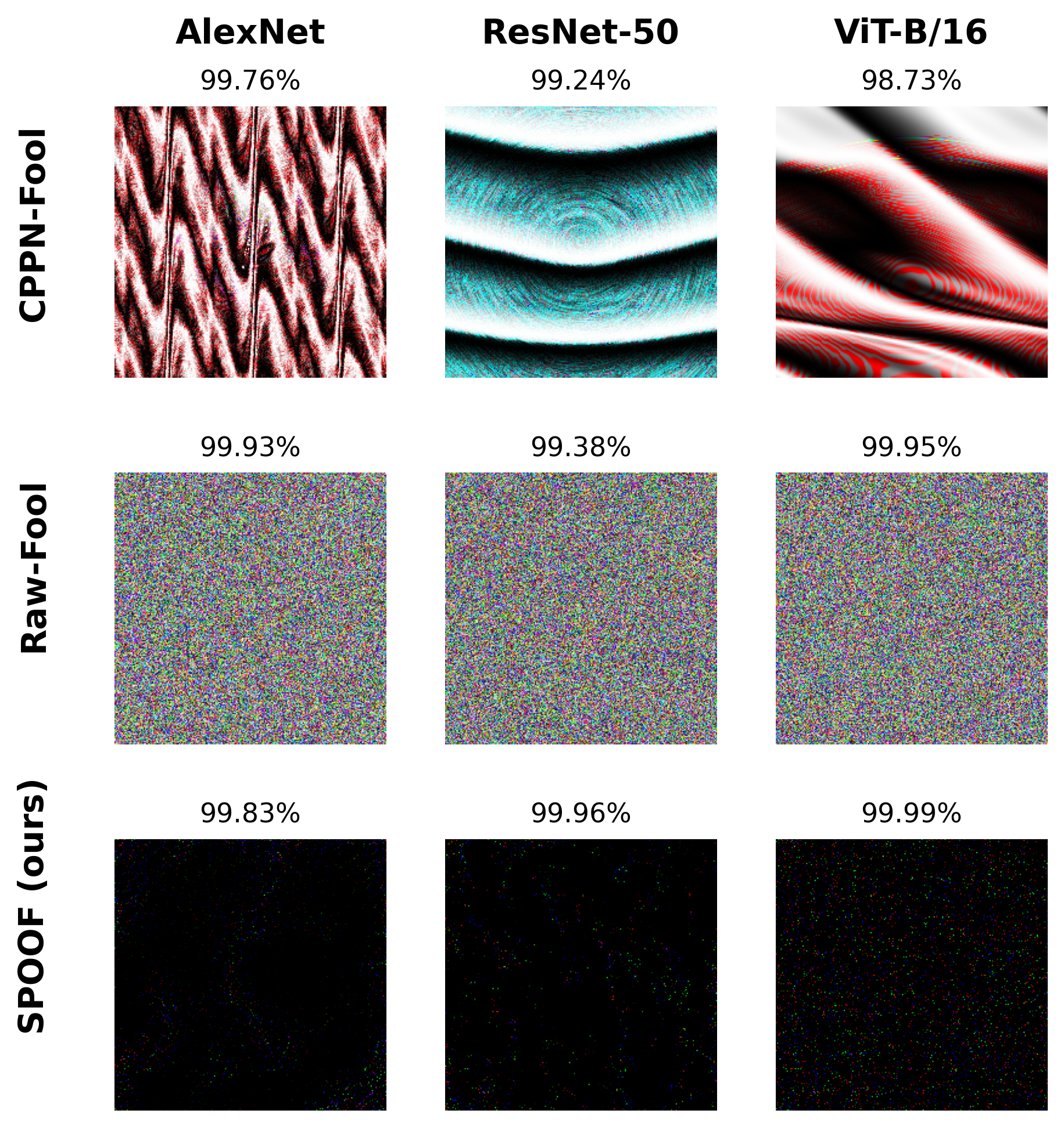}
\caption{Fooling images generated for \textbf{Class 885 — Velvet Fabric}.}
\label{fig:fooling_885}
\end{figure*}

\begin{figure*}[t]
\centering
\includegraphics[width=\textwidth]{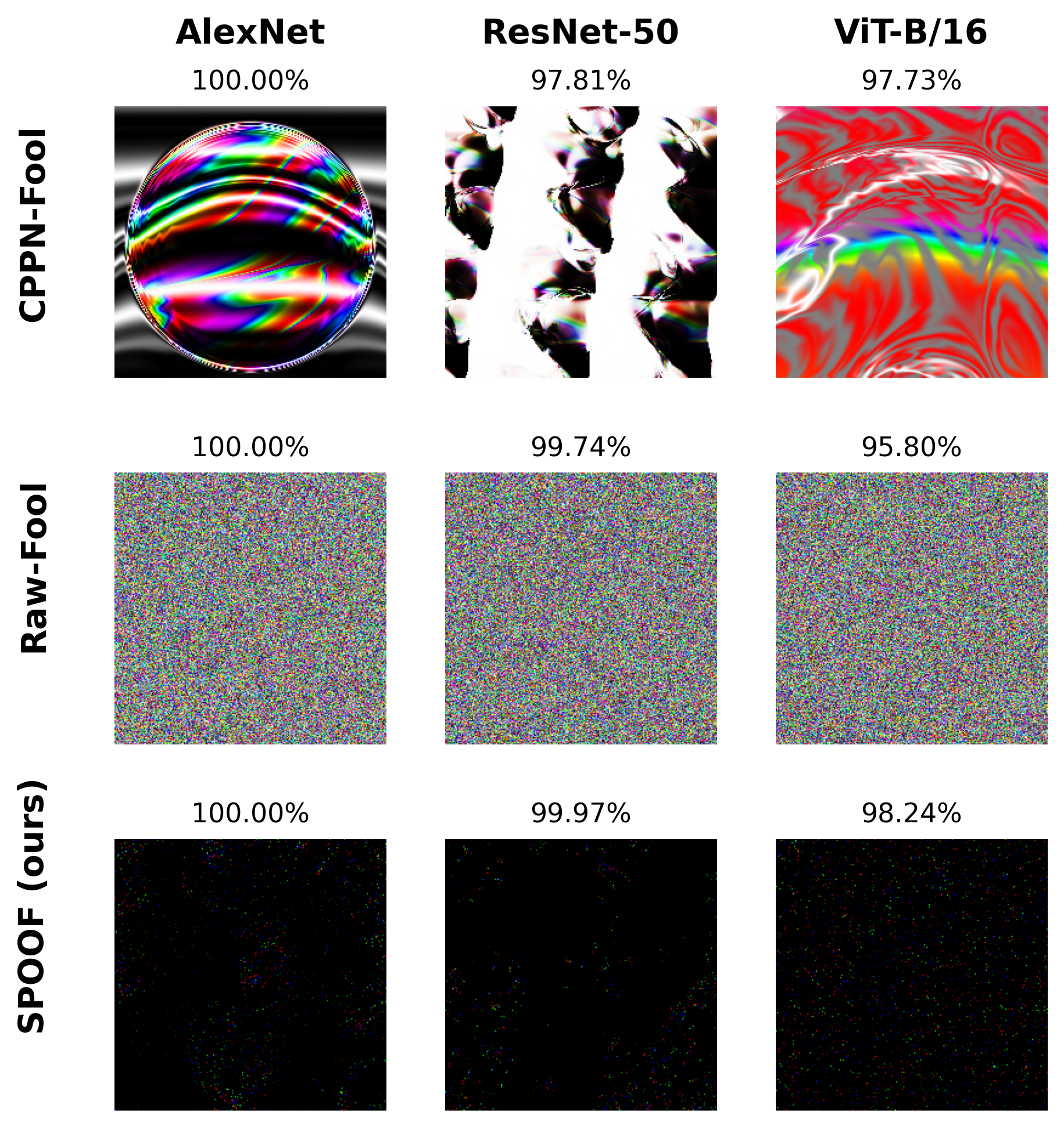}
\caption{Fooling images generated for \textbf{Class 971 — Bubble}.}
\label{fig:fooling_971}
\end{figure*}

\subsection*{S1.2: Mean Aggregated Metrics}
The main paper reports median confidence and PCR across 1000 ImageNet target classes. Here, we include the corresponding mean metrics, offering a complementary aggregation that further characterizes the distribution of fooling performance for each attack–classifier pair (see \cref{tab:mean_results,tab:mean_retrained_results,fig:mean_cppn,fig:mean_raw,fig:mean_spoof})


\section*{S2: MNIST Results on \methodName}
\begin{figure*}[t]
\centering
\includegraphics[width=\textwidth]{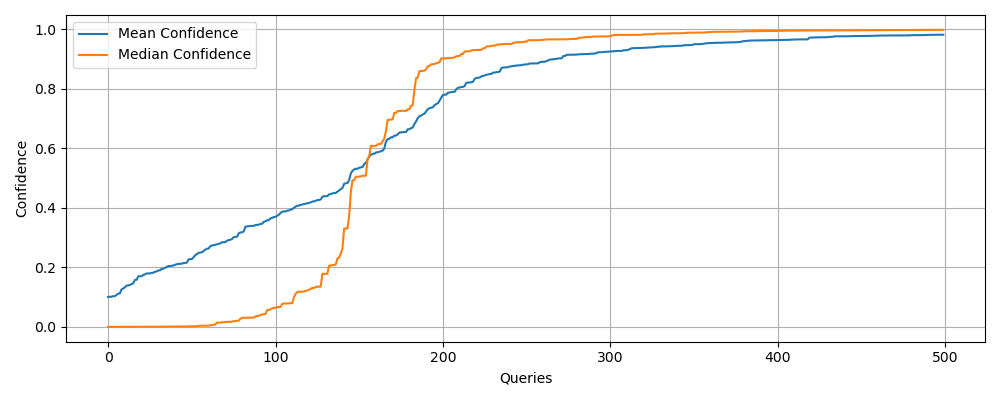}
\caption{Evolution of mean and median SPOOF confidence over 500 queries on MNIST.}
\label{fig:mnist_conf_plot}
\end{figure*}

\begin{figure*}[t]
\centering
\includegraphics[width=\textwidth]{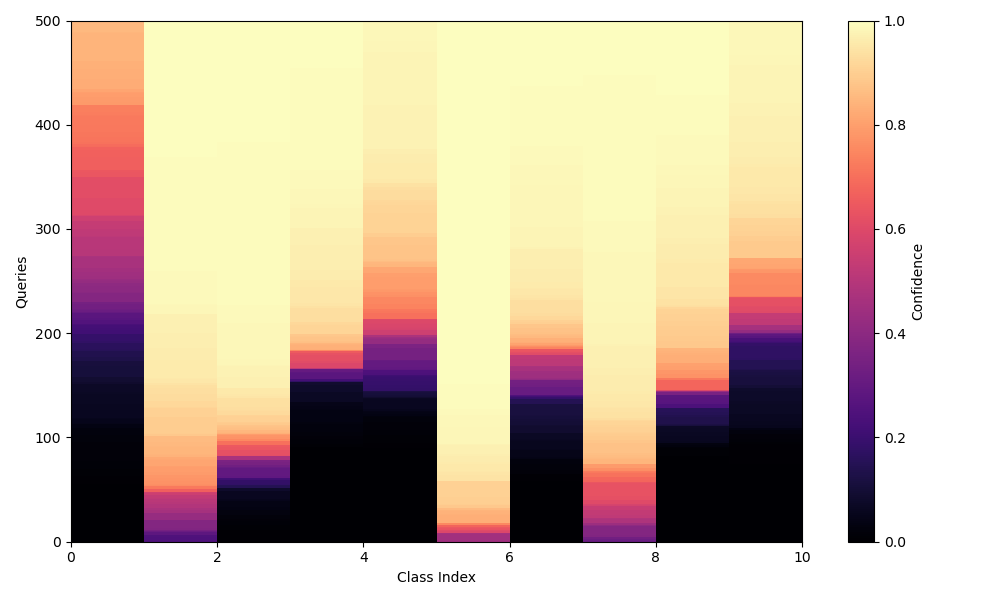}
\caption{Confidence heatmap showing the per-class confidence trajectory across all queries on MNIST.}
\label{fig:mnist_conf_heatmap}
\end{figure*}

\begin{figure*}[t]
\centering
\setlength{\tabcolsep}{4pt}
\renewcommand{\arraystretch}{1}

\begin{tabular}{cccccccccc}
\includegraphics[width=0.085\textwidth]{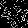} &
\includegraphics[width=0.085\textwidth]{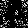} &
\includegraphics[width=0.085\textwidth]{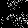} &
\includegraphics[width=0.085\textwidth]{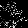} &
\includegraphics[width=0.085\textwidth]{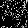} &
\includegraphics[width=0.085\textwidth]{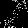} &
\includegraphics[width=0.085\textwidth]{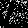} &
\includegraphics[width=0.085\textwidth]{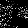} &
\includegraphics[width=0.085\textwidth]{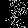} &
\includegraphics[width=0.085\textwidth]{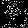} \\

\small{Class 0} & \small{Class 1} & \small{Class 2} & \small{Class 3} & 
\small{Class 4} & \small{Class 5} & \small{Class 6} & \small{Class 7} &
\small{Class 8} & \small{Class 9}
\end{tabular}

\caption{SPOOF-generated MNIST fooling images for all ten target classes.}
\label{fig:mnist_final_images}
\end{figure*}

We include MNIST experiments to show that the behaviors observed on ImageNet—such as sparse fooling and relative classifier vulnerability—also manifest on a second dataset. See \cref{fig:mnist_conf_heatmap,fig:mnist_conf_plot} for confidence plots and \cref{fig:mnist_final_images} for fooling images generated for MNIST trained LeNet.

\end{document}